\documentclass[10pt,twocolumn,letterpaper]{article}

\usepackage{cvpr}
\usepackage{times}
\usepackage{epsfig}
\usepackage{graphicx}
\usepackage{amsmath}
\usepackage{amssymb}
\usepackage{algorithm}
\usepackage{algorithmic}
\usepackage[czech,english]{babel}

\cvprfinalcopy 

\def\cvprPaperID{2167} 

\ifcvprfinal\fi
\begin{document}
\title{Discrete Optimization of Ray Potentials for Semantic 3D Reconstruction}

\author{Nikolay Savinov$^1$, \v Lubor Ladick\'y$^1$, Christian H\"ane and Marc Pollefeys\\
ETH Z\"urich, Switzerland\\
{\tt\small \{nikolay.savinov,lubor.ladicky,christian.haene,marc.pollefeys\}@inf.ethz.ch}
}

\maketitle

\begin{abstract}
Dense semantic 3D reconstruction is typically formulated as a discrete or continuous problem over label assignments in a voxel grid, combining semantic and depth likelihoods in a Markov Random Field framework. The depth and semantic information is incorporated as a unary potential, smoothed by a pairwise regularizer. However, modelling likelihoods as a unary potential does not model the problem correctly leading to various undesirable visibility artifacts.

We propose to formulate an optimization problem that directly optimizes the reprojection error of the 3D model with respect to the image estimates, which corresponds to the optimization over rays, where the cost function depends on the semantic class and depth of the first occupied voxel along the ray. The 2-label formulation is made feasible by transforming it into a graph-representable form under QPBO relaxation, solvable using graph cut. The multi-label problem is solved by applying $\alpha$-expansion using the same relaxation in each expansion move.
Our method was indeed shown to be feasible in practice, running comparably fast to the competing methods, while not suffering from ray potential approximation artifacts.
\end{abstract}

\footnotetext[1]{The authors assert equal contribution and joint first authorship}

\section{Introduction}

In this paper we are studying the problem of jointly inferring dense 3D geometry and semantic labels from multiple images, formulated as an optimization over rays. The problem of dense 3D reconstruction from images and semantic segmentation is up to date still a hard problem. A particularly powerful approach to the problem of dense 3D reconstruction from images is to pose it as a volumetric labeling problem. The volume is segmented into occupied and free space (the inside and the outside of an object) and the surface is extracted as the boundary in between. Traditionally, the data costs are extracted from the input images either directly by computing matching scores per voxel or by first computing depth maps and deriving a per pixel unary potential based on the depth maps. In both cases after the image data has been converted to a per voxel unary term the input images are discarded. Unary terms, approximately modelling the likelihood the depth for a given pixel agrees with the estimate, typically encourage voxels in an interval just before the matched 3D point to take the free-space label and voxels in an interval right after the matched 3D point to take the foreground label. However, this assumption does not hold in general. The interval right behind the corner of an object does not necessarily have to belong to foreground. Failures due to this problem lead to blowing up corners, roofs of buildings or thin objects. The problem can be partly fixed by decreasing the length of an interval, however, in that case in presence of noise the matched neighbouring points start to compete against each other. Another problem of unary approximation is that the unary potential does not model, whether the voxel is visible. If there is a hole in the wall and a matched object behind, there is no penalty associated with a data term for closing the hole, which might get filled in by regularization. This problem can also be partially resolved by penalizing foreground for all pixels in front of the matched voxel, however such solution is not robust to outliers. The standard approach is also not suitable for incorporating multiple candidate matches along the viewing ray in the optimization together.

We propose to formulate an optimization problem which measures the data fidelity directly in image space while still having all the benefits of a volumetric representation. The main idea is to use a volumetric representation, but describe the data cost as a potential over rays. Traversing along a ray from the camera center we observe free space until we first hit an occupied voxel of a certain semantic class and we cannot assume anything about the unobserved space behind. The potential we introduce correctly assigns for each ray the cost, based on the depth and semantic class of the first occupied voxel along the ray. The key to make such a formulation feasible is the transformation of the potential into a graph-representable form under QPBO relaxation~\cite{BorosH02}, solvable using graphcut-based methods. Our proposed optimization method can be directly used also for multi-label problems by applying $\alpha$-expansion~\cite{Boykov01fastapproximate} with QPBO used to calculate each expansion move. Our method runs comparably fast to the competing methods, while not suffering from ray potentials approximation artifacts.

\subsection{Related Work}

Generating dense 3D models out of multiple images is a well-studied problem in computer vision. An overview is given in~\cite{seitz2006comparison}. Posing the problem of dense 3D reconstruction as a volumetric segmentation was first proposed in~\cite{curless1996volumetric}. The initial formulation does not use any regularization. However, often the data is contaminated by noise and strong assumptions about the smoothness have to be made. Regularizing the surface by penalizing the surface area has been proposed in the discrete graph cut formulation~\cite{lempitsky2007global,vogiatzis2005multi} and also as continuous convex optimization formulation~\cite{zach2008fast}. The solution space is not restricted to regular voxel grids. \cite{Vu2012} uses a thetrahedronization of the space. The data cost is formulated as a pairwise potential along viewing rays. They put a unary prior for a tetrahedron right behind the initial estimated depth match and a penalty for every cut before it. The photo-consistency of faces is used as a weight for the cost to cut a ray with a given face. A Markov Random Field (MRF) formulation over rays has been proposed to estimate surface geometry and appearance jointly~\cite{liu2010ray,liu2011complete}. The energy is formulated as a reprojection error of the reconstructed voxels to the input images, which jointly estimates voxel occupancy and color. The high quality refining of approximate mesh has also been formulated as a reprojection error minimization~\cite{vu2009mvs,delaunoy2014photometric}.

The silhouettes of objects in the input image contain important information about the geometry. They constrain the solution, such that for every ray passing through the silhouette there must be at least one occupied voxel, and every ray outside of the silhouette consists of free space voxels only. This constraint has been used in form of a convex relaxation in \cite{cremers2011multiview}. \cite{Hernandez07} proposes an intelligent unary ballooning visibility term based on the consensus from different views. In \cite{Sinha07}, the silhouettes are handled in a two-stage process, where the initial surface is reprojected into each image and the interior is heuristically corrected using the sets of erroneous pixels, by finding the most photo-consistent voxels along the ray. Recently, also approaches which jointly reason about geometry and semantic labels have been proposed~\cite{Hane13}. For volumetric 3D reconstruction in indoor environments a Conditional Random Field (CRF) model was proposed in \cite{kim20133d}. It includes higher order potentials over groups of voxels to include priors from 2D object detections and 3D surface detections in the raw input depth data. Furthermore, potentials over rays are used to enforce visibility of only one voxel along a ray.

\section{Ray Energy Formulation}

We are interested in finding the smooth solution, whose projection into each camera agrees with the depth and semantic observations. Thus, the energy will take the form:
\begin{equation}
E({\bf x}) = \sum_{r \in {\cal R}} \psi_r({\bf x}^r) + \sum_{(i, j) \in {\cal E}} \psi_p(x_i, x_j),
\end{equation}
where each $x_i \in {\cal L}$ is the voxel variable taking a label from the label set ${\cal L}$ with a special label $l_f \in {\cal L}$ corresponding to free space;  ${\cal R}$ is the set of rays, $\psi_r(.)$ is the ray potential over the set of voxels ${\bf x}^r$, ${\cal E}$ is the set of local voxel neighbourhoods, and $\psi_p(\cdot)$ is a smoothness enforcing pairwise regularizer. Each ray $r$ of length $N_r$ consists of voxels $x^r_i = x_{r_i}$, where $i \in \{0, 1, .. N_r - 1\}$. The ray potential takes the cost depending only on the first non-free space voxel along the ray $K^r$ (if there is any):
\begin{equation}\label{ray_first}
K^r =
 \begin{cases} \min(i | x^r_i \neq l_f) & \text{if } \exists x^r_i \neq l_f\\
N_r & \text{otherwise.}
  \end{cases}
\end{equation}
The ray potential is defined as:
\begin{equation}\label{ray_pot}
\psi_r({\bf x}^r) = \phi_r(K^r, x^r_{K^r}),
\end{equation}
where $x^r_{N^r} = l_f$. The costs for each depth and semantic label $\phi_r(.)$ could be arbitrary and typically come from the corresponding semantic and depth classifiers or measurements.

\subsection{Optimization of the 2-label Ray Potentials}
Let us first consider the two label case. Each variable $x_i$ may take a value from the set $\{0, 1\}$, where $x_i = 0$ corresponds to
the occupied voxel and $x_i = 1$ corresponds to free space $l_f$. The ray potential (\ref{ray_pot}) for arbitrary costs could be non-submodular even for a ray of the length $2$ making the 2-label problem NP-hard in general.
Thus, we propose a solution using QPBO relaxation~\cite{BorosH02}, where the energy $E({\bf x})$ is transformed to a submodular energy $E({\bf x}, \overline{\bf x})$ with additional variables $\overline{x_i} = 1 - x_i$, and solved by relaxing these constraints.

In our case, the non-submodular ray potentials are of larger order than 2. To make our problem solvable using graph-cut, our goal is to transform these potentials into a pairwise energy with additional auxiliary variables ${\bf z}$, such that:
\begin{equation}
\psi_r({\bf x}^r) = \min_{\bf z} \psi_q({\bf x}^r, {\overline{\bf x}^r}, {\bf z}),
\end{equation}
where $\psi_q(.)$ is pairwise submodular. Additionally, to keep the problem feasible, our goal is to find a transformation, for which the number of edges in the graph with auxiliary variables grows at most linearly with length of a ray. We achieve this goal using these five steps:
\begin{enumerate}
\item Polynomial representation of the ray potential,
\item Transformation into higher order submodular potential using additional variables $\overline{\bf x}$,
\item Pairwise graph construction of a higher order submodular potential using auxiliary variables ${\bf z}$,
\item Merging variables~\cite{ramalingam2011efficient} to get the linear dependency of the number of edges on length,
\item Transformation into a normal form, symmetric over ${\bf x}$ and $\overline{\bf x}$, suitable for QPBO~\cite{BorosH02}.
\end{enumerate}

Next we describe in details each one of these steps. The two-label equivalent of the ray potential takes the form:
\begin{equation}\label{ray_pot2}
\psi_r({\bf x}^r) :=
 \begin{cases} \phi_r(\min(i | x^r_i = 0)) & \text{if } \exists x^r_i = 0\\
\phi_r(N_r) & \text{otherwise,}
  \end{cases}
\end{equation}
where $\phi_r(i) := \phi_r(i)$ is the cost taken, if $i$ is the first foreground pixel along the ray, and $\phi_r(N_r)$ is the cost for the whole ray being free space.
We would like to transform this potential into the polynomial representation - the sum of products:
\begin{equation}
\psi_r({\bf x}^r) = k^r + \sum_{i = 0}^{N_r-1} c^r_i \prod_{j = 0}^i x^r_j.
\end{equation}
Applying to equation (\ref{ray_pot2}), we get $\phi_r(K) = k^r + \sum_{i = 0}^{K-1} c^r_i$, thus $k^r = \phi_r(0)$ and $c^r_{i} = \phi_r(i+1) - \phi_r(i), \text{ } \forall i \in \{0, .. N_r - 1\}$.

It is well-known that the product $c^r_i \prod_{j = 0}^i x^r_j$ is submodular only if $c^r_i \leq 0$~\cite{FreedmanD05}. For $c^r_i > 0$ we can transform the product into submodular function using additional variable $\overline{x}^r_i = 1 - x^r_i$ as:
\begin{equation}
c^r_i \prod_{j = 0}^i x^r_j = c^r_i (1 - \overline{x}^r_i) \prod_{j = 0}^{i - 1} x^r_j = -c^r_i \overline{x}^r_i \prod_{j = 0}^{i - 1} x^r_j + c^r_i \prod_{j = 0}^{i - 1} x^r_j\label{posterm}.
\end{equation}
That means, that starting from the last term we can iteratively check if $c^r_i \leq 0$, and if it is not, we transform the term using (\ref{posterm}) and update $c^r_{i - 1} := c^r_{i - 1} + c^r_i$. The transformation algorithm is explained in details in Algorithm~\ref{transf_sub}.
\begin{algorithm}
   \caption{{\it Transformation into submodular potential.}}
   \label{transf_sub}
\begin{algorithmic}
   \STATE {\bfseries Input:} ${\bf c}^r$, $N_r$
   \STATE {\bfseries Output:} ${\bf a}^r$, ${\bf b}^r$

   \STATE {$i = N_r - 1;$}\\
   \WHILE {$i \geq 0$}
       \IF{$c^r_i \leq 0$}
       \STATE $a^r_i = -c^r_i$, $b^r_i = 0$\\
       \ELSE
           \STATE $a^r_i = 0$, $b^r_i = c^r_i$\\
           \textbf{if} $i > 0$ \textbf{then} $c^r_{i - 1} = c^r_{i - 1} + c^r_i$\\
       \ENDIF
       \STATE $i = i - 1$\\
   \ENDWHILE
\end{algorithmic}
\end{algorithm}

Ignoring the constant term we transform the potential into:
\begin{equation}
\psi_r({\bf x}^r) = \sum_{i = 0}^{N_r - 1} \bigg{(}-a^r_i \prod_{j = 0}^i x^r_j - b^r_i \overline{x}^r_i \prod_{j = 0}^{i-1} x^r_j \bigg{)}.
\end{equation}
Each product in the sum is submodular and graph-representable using one auxiliary variable $z_i \in \{0,1\}$. The standard pairwise graph constructions~\cite{FreedmanD05} for a negative product terms are:
\begin{equation}
-a^r_i \prod_{j = 0}^i x^r_j = a^r_i \min_{z_i} \bigg{(} -z_i + \sum_{j = 0}^i z_i(1 -  x^r_j) \bigg{)}
\end{equation}
\begin{eqnarray}
-b^r_i \overline{x}^r_i \prod_{j = 0}^{i - 1} x^r_j &=& b^r_i \min_{z_i'} \bigg{(} -z_i' + z_i' (1 - \overline{x}^r_i) \nonumber\\
&+& \sum_{j = 0}^{i - 1} z_i'(1 -  x^r_j) \bigg{)},
\end{eqnarray}
however, such constructions for each term in the sum would lead to a quadratic number of edges per ray. Instead, we first build a more complex graph constructions with $(i+1)$ auxiliary variables ${\bf z}^i \in \{0,1\}^{i+1}$ and ${\bf z'}^i \in \{0,1\}^{i+1}$ respectively with the foresight, that this will lead to a graph construction with linear growth of the number of edges:
\begin{eqnarray}
-a^r_i\prod_{j = 0}^i x^r_j &=& a^r_i \min_{{\bf z}_i} \bigg{(} -z_i^i + z_i^i (1 - x^r_i)\label{pos_eq}\\
 &+& \sum_{j = 0}^{i - 1} (z_{j + 1}^i (1 - z_j^i) + z_j^i(1 - x^r_j)) \bigg{)}\nonumber
\end{eqnarray}
\begin{eqnarray}
-b^r_i \overline{x}^r_i \prod_{j = 0}^{i - 1} x^r_j &=& b^r_i \min_{{\bf z'}_i} \bigg{(} -{z'}_i^i +  {z'}_i^i (1 - \overline{x}^r_i) \label{neg_eq}\\
 &+& \sum_{j = 0}^{i - 1} ({z'}_{j + 1}^i (1 - z'^j_i) + {z'}^i_j(1 - x^r_j)) \bigg{)},\nonumber
\end{eqnarray}
where the (one of the) optimal assignment for auxiliary variables is:
\begin{eqnarray}
\forall j \in \{0, .., i\} : z_j^i &=& \prod_{k = 0}^j x^r_k\label{assign1}\\
{z'}_i^i &=& \overline{x}^r_i \prod_{k = 0}^{i-1} x^r_k\label{assign2}\\
\forall j \in \{0, .., i-1\} : {z'}_j^i &=& \prod_{k = 0}^j x^r_k\label{assign3}
\end{eqnarray}
Both equations are structurally the same, so we just prove (\ref{pos_eq}). It can be easily checked, that given the assignments of auxiliary variables (\ref{assign1}) every term is always going to be $0$ except $-z_i^i$, that is by definition (\ref{assign1}) equal to the negative product of all ${\bf x}^r$ on the left side of (\ref{pos_eq}). Next we have to show, that if ${\bf x}^r \neq {\bf 1}$, there is no assignment of auxiliary variables leading to a negative cost. Let us try to construct such ${\bf x}^r$. The only term, that could be potentially negative is $-z^i_i$, thus $z^i_i = 1$. However, to keep the whole expression negative, the remaining terms must be $0$ and thus:
\begin{eqnarray}
(z_{j+1}^i (1-z_j^i) = 0) \implies ((z^i_{j+1} = 1) \implies (z_j^i = 1))\label{all_zs}\\
(z_j^i (1-x_i^r) = 0) \implies ((z^i_j = 1) \implies (x^r_j = 1))\label{all_xs}.
\end{eqnarray}
(\ref{all_zs}) implies ${\bf z}^i = {\bf 1}$ and then $(\ref{all_xs})$ implies ${\bf x}^r = {\bf 1}$. Thus, the assignment (\ref{assign1}) is always optimal and (\ref{pos_eq}) holds. There could be more possible optimal assignments, however for the next step we only need this one to exist.
\begin{figure*}[t]
\centering
\includegraphics[width=1\linewidth]{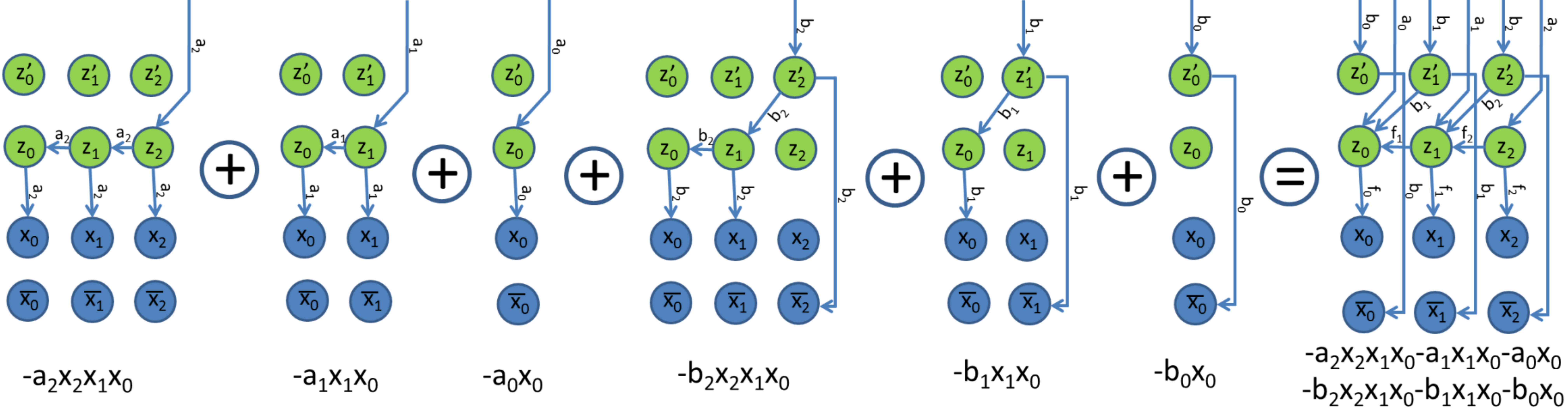}
\caption{The process of merging for ray of the length 3. Because the optimal assignment of auxiliary variables is the same for every construction, the variables can be merged and thus we can build the final graph just by summing up corresponding edges in each one of these $6$ graphs. The weights in the final graph will be $f_0 = a_2 + a_1 + a_0 + b_2 + b_1$, $f_1 = a_2 + a_1 + b_2$ and $f_2 = a_2$.
\label{graph_merge}}
\end{figure*}
The alternative graph construction leads to much more complex graphs than the standard one, however, we can decrease its growth in terms of edges by applying the merging theorem~\cite{ramalingam2011efficient}. The theorem states, that if for every assignment of input variables ${\bf x}$ there exists at least one assignment of two or more auxiliary variables $z_i, .. z_j$, such that $z_i = .. = z_j$, these variables can be replaced by a single variable without altering the cost for any assignment of ${\bf x}$. In our graph constructions we can see ($\ref{assign1}$), that the value $z_j^i = \prod_{k = 0}^j x^r_k$ is independent on $i$, and thus:
\begin{equation}
 \forall i \in \{0, .., N_r - 1\}, \forall j,k \in \{0, .., i\} : z^j_i = z^k_i.
\end{equation}
Furthermore, based on ($\ref{assign3}$):
\begin{equation}
\forall i \in \{0, .., N_r - 1\}, \forall j,k \in \{0, .., i - 1\} : {z'}^j_i = z^k_i.
\end{equation}
After removing the unnecessary top index $z^j_i = z_i$ and ${z'}^i_i = z'_i$, and summing up all corresponding weights, the resulting pairwise construction takes the form:
\begin{eqnarray}
\psi_r({\bf x}^r) &=& \min_{{\bf z}, {\bf z'}} \Bigg{(} \sum_{i = 0}^{N_r - 1} \bigg{(}-a^r_i z_i - b^r_i z_i'\\
 &+& f^r_i (1 - z_i) x^r_i + b^r_i (1 - z_i) \overline{x}^r_i \bigg{)}\nonumber\\
 &+& \sum_{i = 1}^{N_r - 2} \bigg{(}(f^r_{i + 1} (1 - z_{i + 1}) z_i + b^r_i (1 - z'_{i + 1}) z_i\bigg{)}\Bigg{)}\nonumber,
\end{eqnarray}
where $f^r_i = \sum_{i}^{N_r-1} a^r_j + \sum_{i + 1}^{N_r-1} b^r_j$. Unlike standard graph construction, this one leads to a number of edges growing at most linearly with the length of the ray. The process of merging is visually depicted in Figure~\ref{graph_merge}.

\begin{figure}
\centering
\includegraphics[width=1\linewidth]{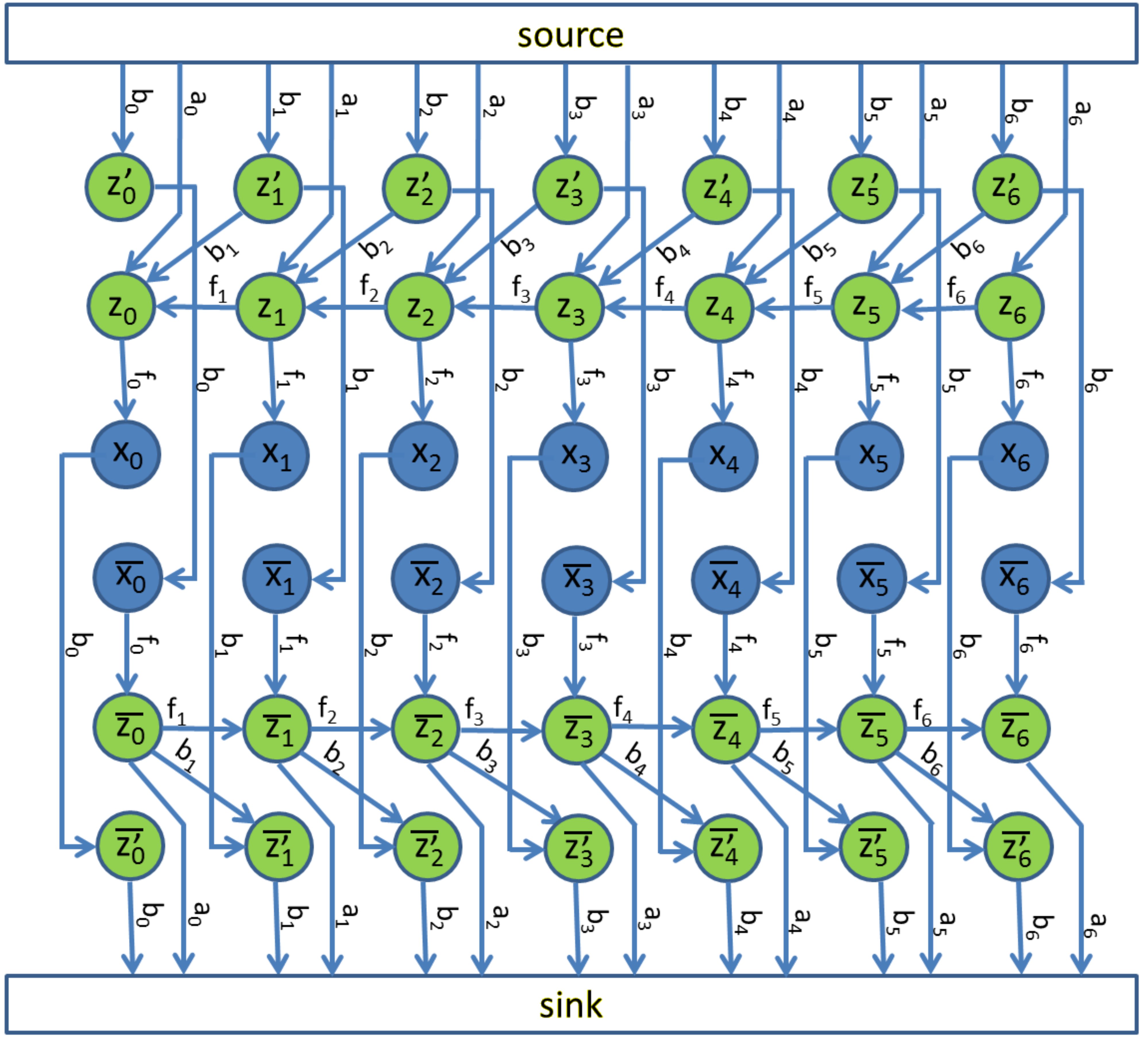}
\caption{The graph construction for ray with 7 voxels. Each variable has at most 3 outgoing edges, thus the total number of them grows linearly with the length of the ray.
\label{graph_ray}}
\end{figure}

Finally, the potential is converted into a symmetric normal form, suitable for QPBO~\cite{BorosH02} as:
\begin{eqnarray}
\psi_r({\bf x}^r) &=& \frac{1}{2} \bigg{(}\min_{\bf z} \psi^r({\bf x}, \overline{\bf x}, {\bf z})\nonumber\\
&+& \min_{\overline{\bf z}} \psi^r({\bf 1} - \overline{\bf x}, {\bf 1} - {\bf x}, {\bf 1} - \overline{\bf z})\bigg{)},
\end{eqnarray}
leading to the symmetric quadratic representation:
\begin{eqnarray}
\psi_r({\bf x}^r) &=& \frac{1}{2} \min_{{\bf z}, {\bf z'}, \overline{\bf z}, \overline{\bf z'}} \bigg{(}\sum_{i = 0}^{N_r-1} \bigg{(}-a^r_i z_i - b^r_i z_i'\label{gc_final}\\
&+& f^r_i (1 - z_i) x^r_i + b^r_i (1 - z_i) \overline{x}^r_i \bigg{)}\nonumber\\
&+& \sum_{i = 1}^{N_r - 2} \bigg{(}f^r_{i + 1} (1 - z_{i + 1}) z_i + b^r_i (1 - z'_{i + 1}) z_i\bigg{)}\nonumber\\
&+& \sum_{i = 0}^{N_r - 1} \bigg{(}-a^r_i (1 - \overline{z}_i) - b^r_i (1 - \overline{z'}_i)\nonumber\\
&+& f^r_i \overline{z}_i (1 - \overline{x}^r_i) + b^r_i \overline{z}_i (1 - x^r_i) \bigg{)}\nonumber\\
&+& \sum_{i = 1}^{N_r - 2} \bigg{(}f^r_{i + 1} \overline{z}_{i + 1} (1 - \overline{z}_i) + b^r_i \overline{z'}_{i + 1} (1 - \overline{z}_i)\bigg{)}\Bigg{)}\nonumber.
\end{eqnarray}
To remove fractions, we multiply all potentials by a factor of 2. The resulting graph construction is shown in Figure~\ref{graph_ray}.
Our method inherits the properties of the standard QPBO for pairwise graphs~\cite{BorosH02}. The set of variables, for which $x_i = 1 - \overline{x}_i$, is the part of the globally optimal solution. The remaining variables can be labelled using Iterated Conditional Modes (ICM)~\cite{Besag86} iteratively per variable.

\subsection{Optimization of the Multi-label Ray Potentials}
To solve the multi-label case, we adapt the $\alpha$-expansion algorithm. The
moves proposed by $\alpha$-expansion~\cite{Boykov01fastapproximate}, are encoded by a binary transformation vector ${\bf t}$, which encodes, how should the variable $x_i$ change after the move. There are two distinct cases - expansion move of the free space label $l_f$ and expansion move of any other foreground label $l \neq l_f$.

For the free space label we define the transformation function as:
\begin{equation}
T_{l_f}(x_i, t_i)  =\left\{
\begin{array}{ccl}
l_f & \mbox{if} & t_i = 1\\
x_i & \mbox{if} & t_i = 0
\end{array}
\right.
\end{equation}
Next we have to find the 2-label projection of multi-label ray potential (\ref{ray_pot}) under this transformation function $T_{l_f}(.)$. Let $s(r)$ be the ordered set of variables in a ray $r$, which before the move do not take the label $l_f$ : $s(r) = \{r_i | x^r_i \neq l_f\}$. After the move, the first occupied voxel of $r$ will be the first occupied voxel of $s(r)$, and thus the ray potential (\ref{ray_pot}) projects into 2 labels as:
\begin{equation}\label{ray_pot3}
\psi_r({\bf t}^r) =
 \begin{cases} \phi_r(K({\bf t}), x^{s(r)}_{K({\bf t})}) & \text{if } \exists t^{s(r)}_i = 0\\
\phi_r(N_r, l_f) & \text{otherwise,}
  \end{cases}
\end{equation}
where $K({\bf t}) = \min(i \in s(r) | t^{s(r)}_i = 0)$. The projection takes the form of (\ref{ray_pot2}), and thus can be solved using the graph-construction (\ref{gc_final}).

For $l \neq l_f$ we define the transformation function as:
\begin{equation}
T_{l}(x_i, t_i)  = \left\{
\begin{array}{ccl}
l & \mbox{if} & t_i = 0\\
x_i & \mbox{if} & t_i = 1
\end{array}
\right.
\end{equation}
After the move the first non-free space label could be only in front of the first non-free space label of the current solution. Let $s'(r)$ be the ordered set of variables in front of current first occupied voxel $K_r$ : $s'(r) = \{r_i | i \leq K_r\}$ (including $K_r$ if $K_r$ exists). The ray potential will project into:
\begin{equation}\label{ray_pot4}
\psi_r({\bf t}^r) =
 \begin{cases} \phi_r(K({\bf t}), l) & \text{if } \exists t^{s'(r)}_i = 0\\
\phi_r(N_{s'(r)}, x^{s'(r)}_{K_r}) & \text{otherwise.}
  \end{cases}
\end{equation}
The projection also takes the form of (\ref{ray_pot2}) and can be solved using (\ref{gc_final}).

\section{Implementation Details}
As an input our method uses semantic likelihoods, predicted by a pixel-wise context-based classifier from~\cite{hierarchicalcrf}, and depth likelihoods obtained by a plane sweep stereo matching algorithm using zero-mean normalized cross-correlation. In general the ray costs for a label $l$ take the form:
\begin{equation}
\phi_r(i, l) = \bigg{(}(\lambda_{sem} C(l) + \lambda_{dep} C(d(i))\bigg{)} d(i)^2,
\end{equation}
where $C(l)$ is the semantic cost for label $l$ (or sky for free space label $l_f$), $C(d(i))$ is the cost for depth $d(i)$ of a pixel $i$, and $\lambda_{sem}$ and $\lambda_{dep}$ are the weights of both domains. Because each pixel corresponds to patch in a volume, which scales quadratically with distance, thus the costs have to be weighted by a factor of $d(i)^2$ to keep constant ratio between ray potentials and regularization terms. In theory, our method allows for optimization over the whole depth cost volume, however this would require too much memory to store. In practice, a few top matches (in our case we used at most 3) for each pixel contain all the relevant information and all remaining scores are typically random noise. In our experiments the costs for given depth took the form:
\begin{equation}
C(d(i)) =
\begin{cases} w_n(-1 + \frac{|d(i) - d^n_{top}|}{\Delta}) & \text{if } |d(i)-d^n_{top}| \leq \Delta\\
0 & \text{otherwise.}
\end{cases}
\end{equation}
where $w_n$ is the weight of the $n$-th match with the depth $d^n_{top}$, calculated as a ratio of confidence scores of the n-th match with respect to the best one. As a smoothness enforcing pairwise potential we used the discrete approximation~\cite{kolmogorov2005metrics} of the continuous anisotropic pairwise regularizer from~\cite{Hane13}. To deal with the high resolution of the $3D$ we use a coarse-to-fine approximation with 3 subdivision stages. As a graph-cut solver, we eventually used the IBFS algorithm~\cite{Goldberg11} algorithm, which was typically $5-50\times$ faster than commonly used Boykov-Kolmogorov~\cite{Boykov01fastapproximate} algorithm optimized for lattice graphs. The run-time depends not only on the resolution, but also on the number of rays sampled. For example for the Castle dataset with $50$ million voxels and $150$ million rays the optimization took approximately 40 minutes on 48 CPU cores, which is comparable or faster than other methods~\cite{Hane13}. To extract mesh out of voxelized solution we used Marching cubes algorithm~\cite{Lorensen87}. Final models were smoothed using Laplacian smoothing to reduce discretization artifacts.

\section{Experiments}
\begin{figure*}
\begin{center}
\begin{tabular}{cccc}
\vspace{4mm}
\includegraphics[width=0.2\linewidth]{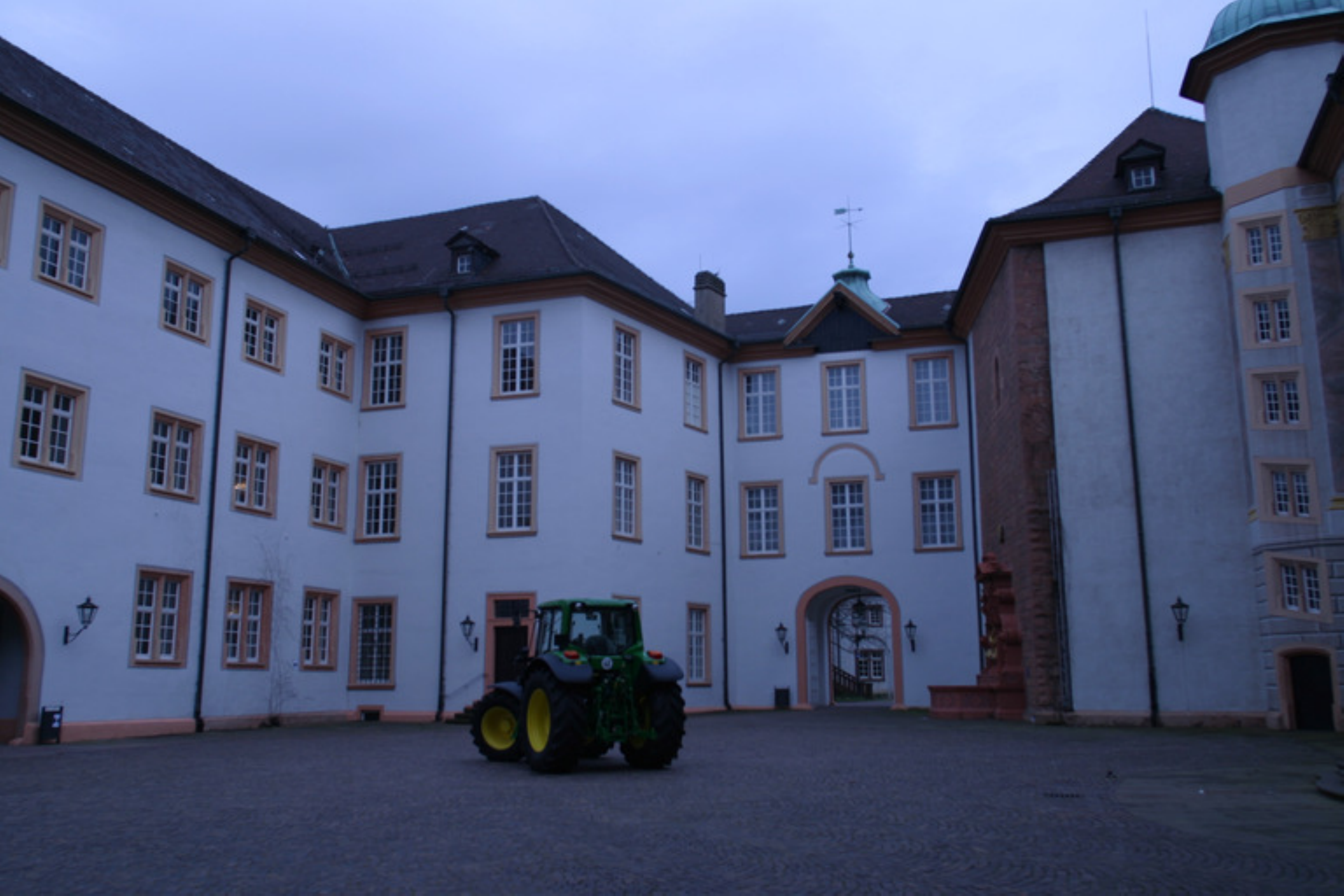}&
\includegraphics[width=0.2\linewidth]{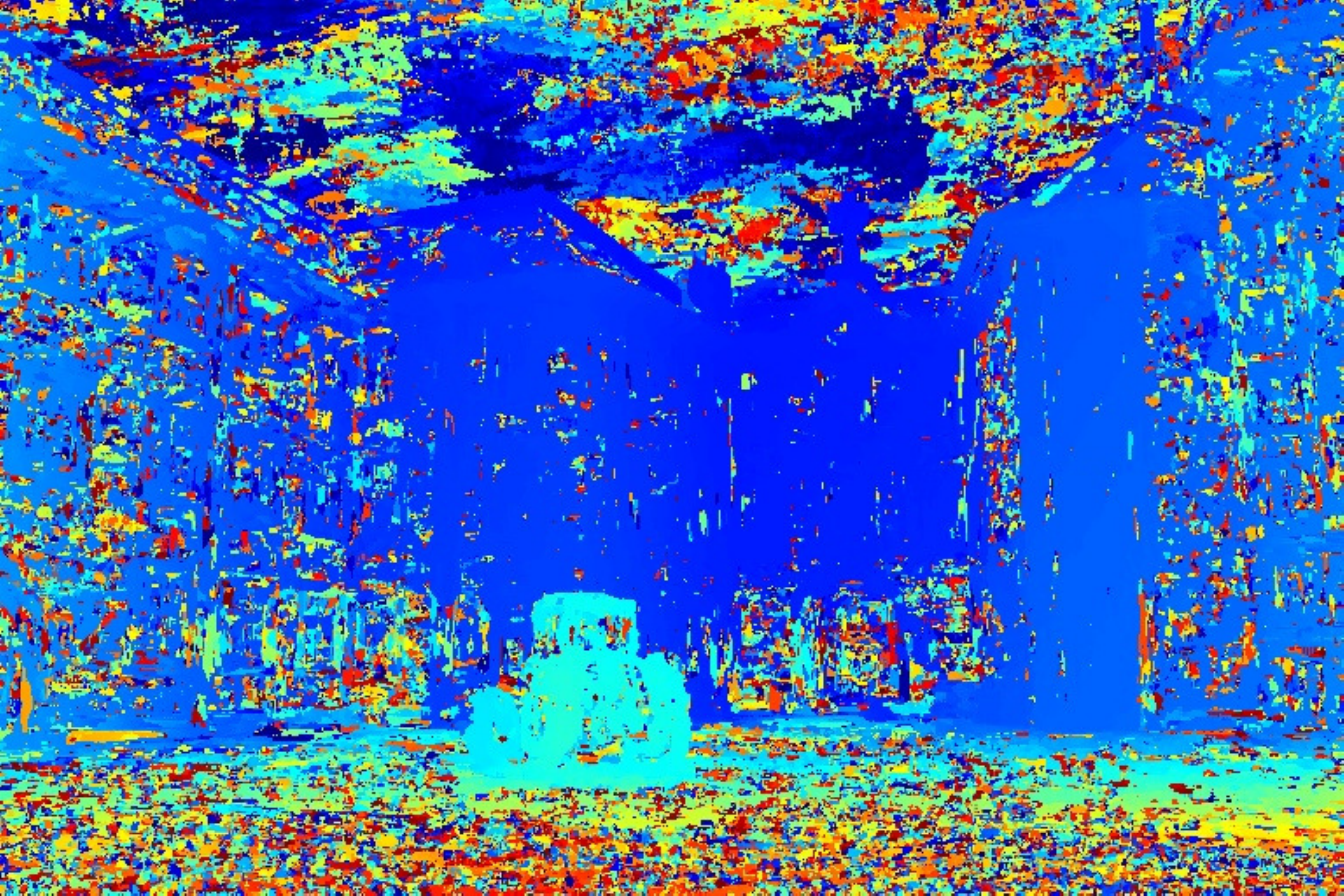}&
\includegraphics[width=0.2\linewidth]{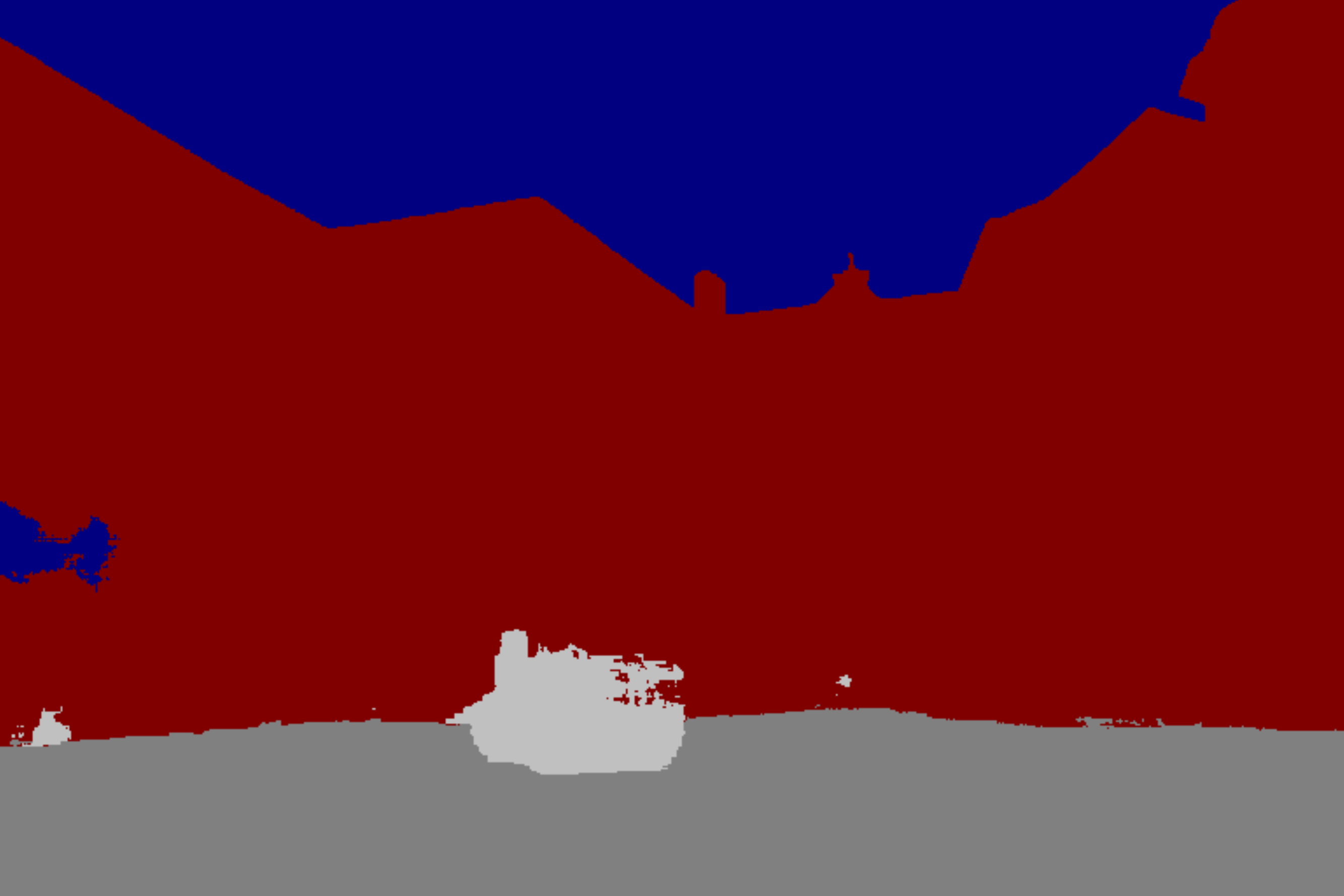}&
\includegraphics[width=0.33\linewidth]{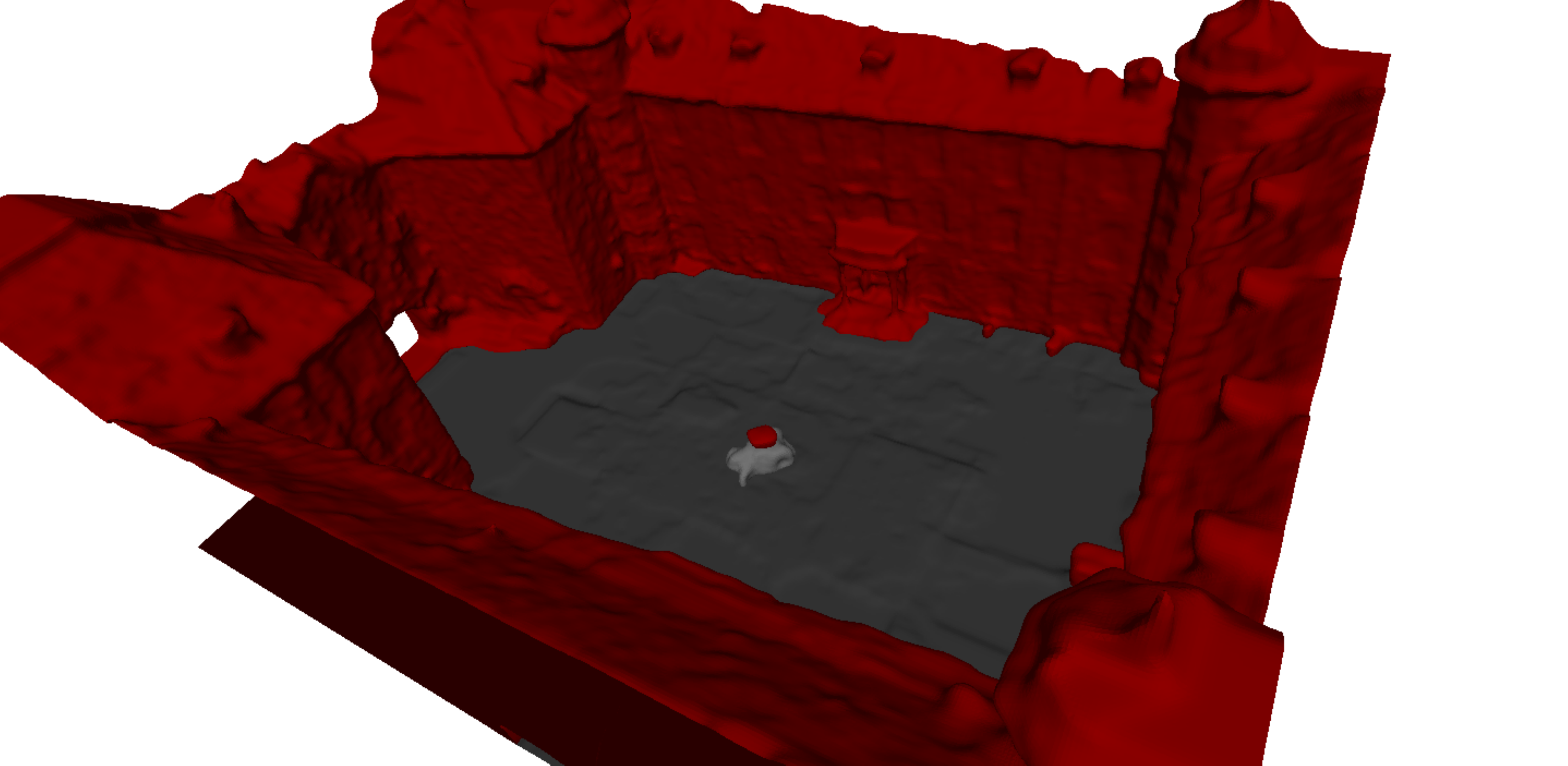}\\
\vspace{4mm}
\includegraphics[width=0.2\linewidth]{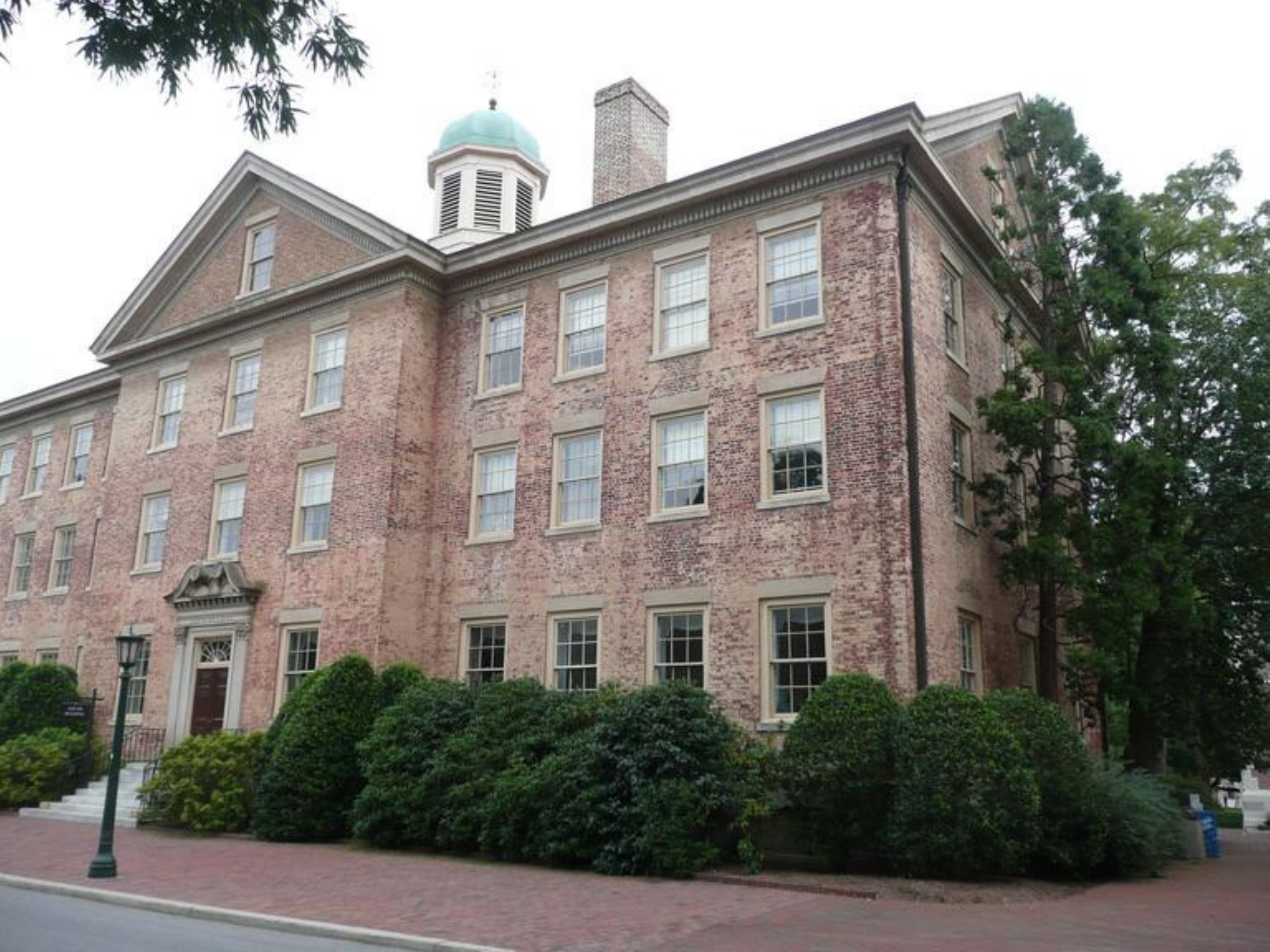}&
\includegraphics[width=0.2\linewidth]{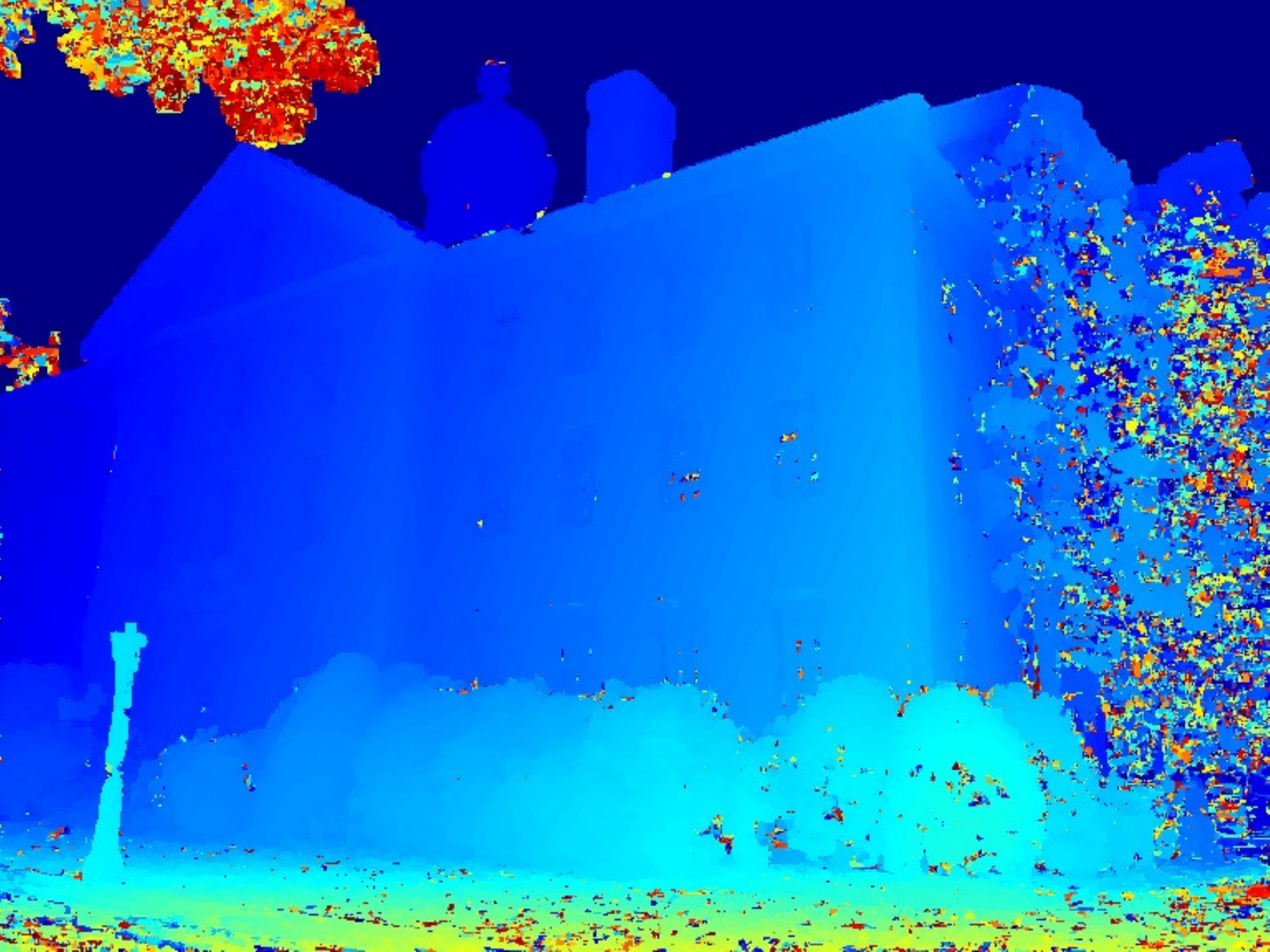}&
\includegraphics[width=0.2\linewidth]{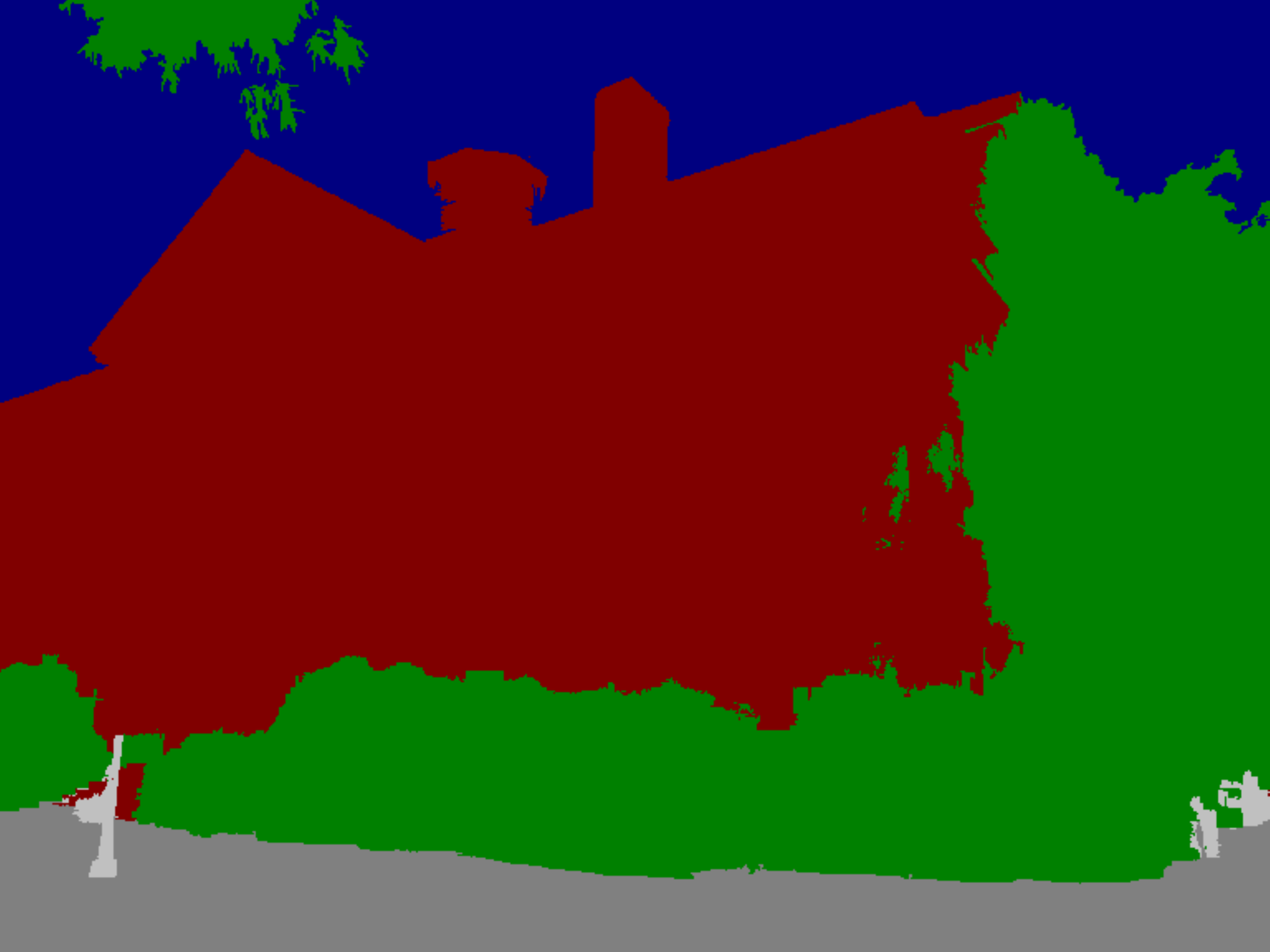}&
\includegraphics[width=0.33\linewidth]{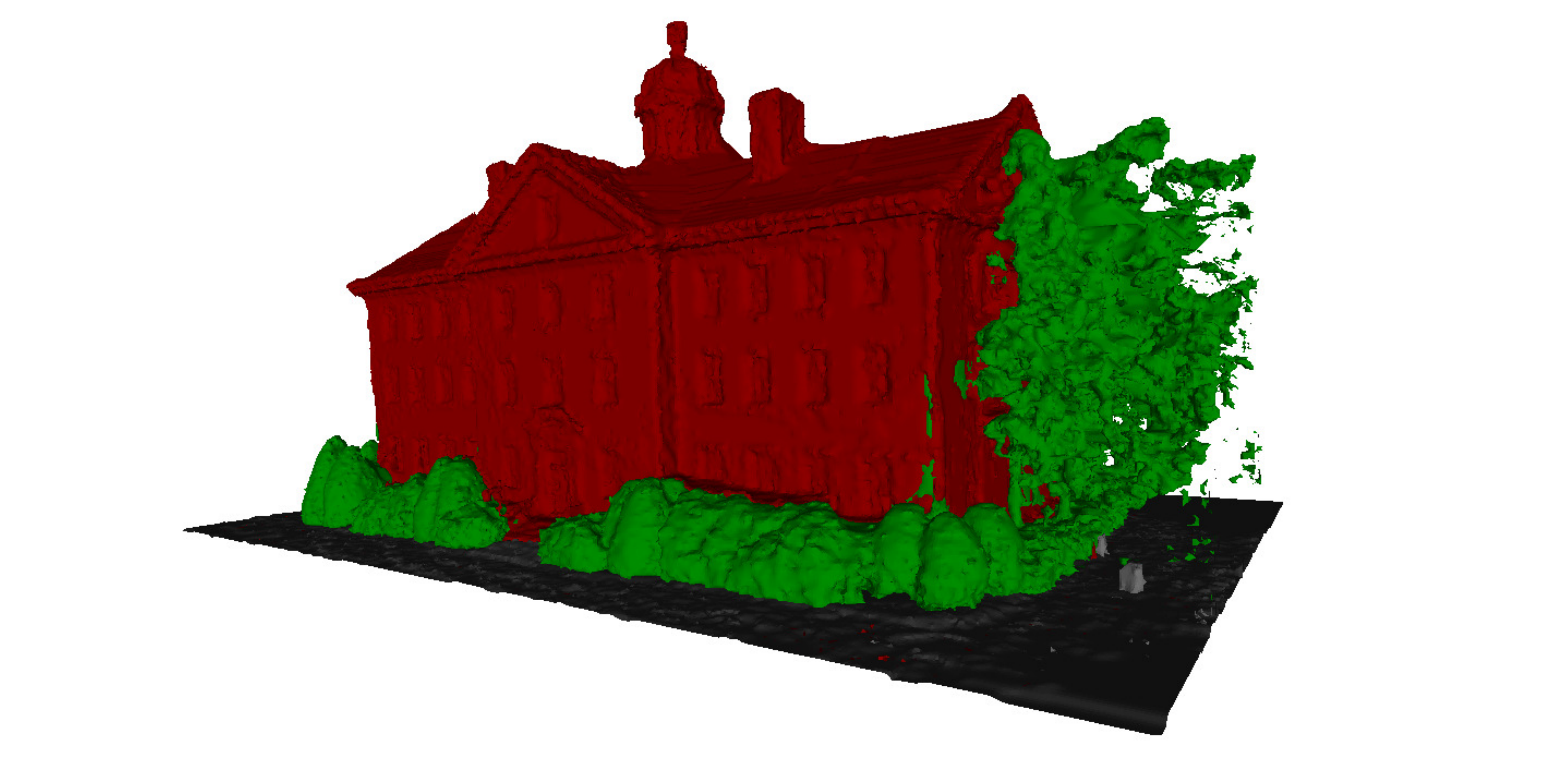}\\
\vspace{4mm}
\includegraphics[width=0.2\linewidth]{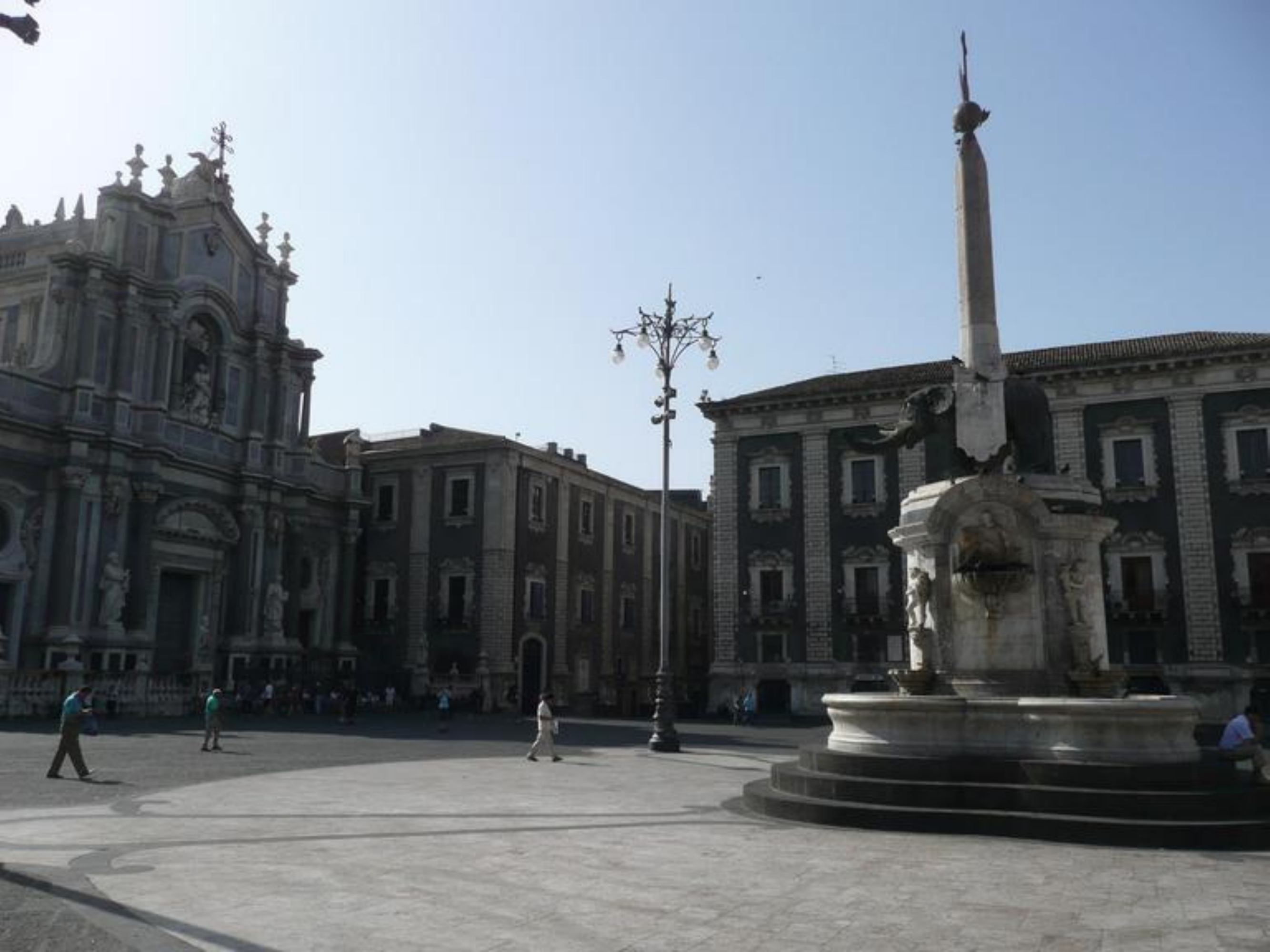}&
\includegraphics[width=0.2\linewidth]{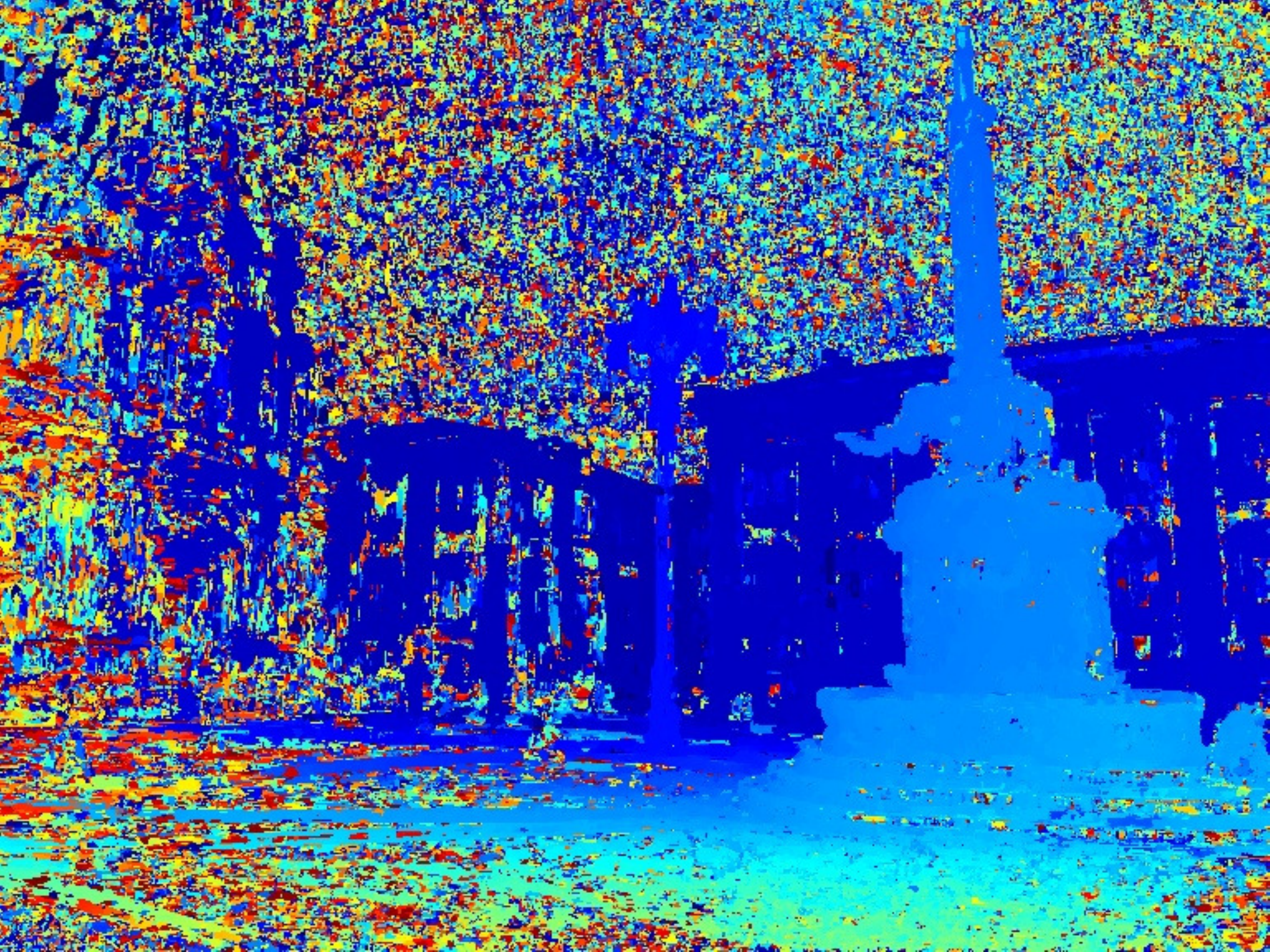}&
\includegraphics[width=0.2\linewidth]{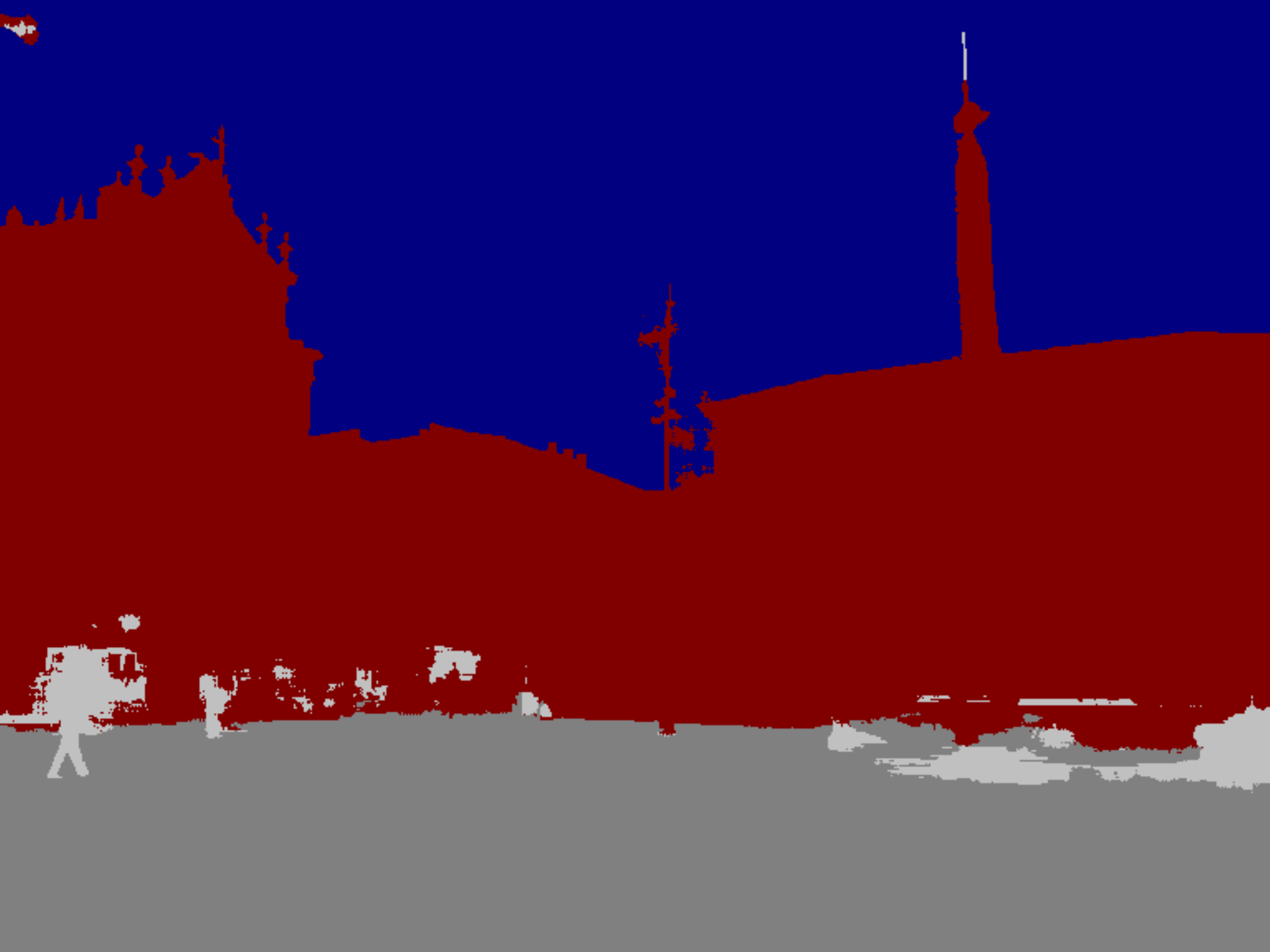}&
\includegraphics[width=0.33\linewidth]{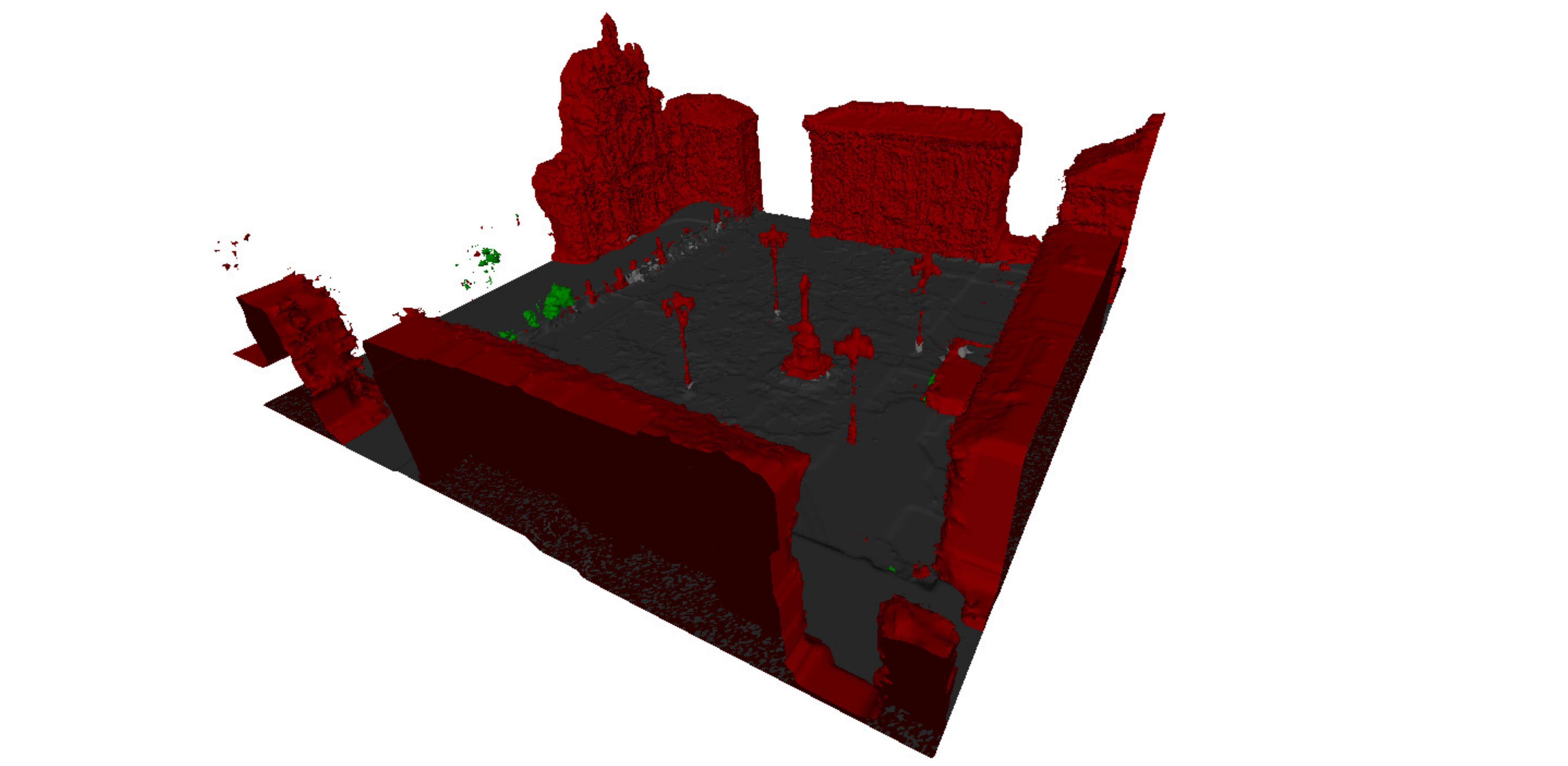}\\
\vspace{4mm}
\includegraphics[width=0.2\linewidth]{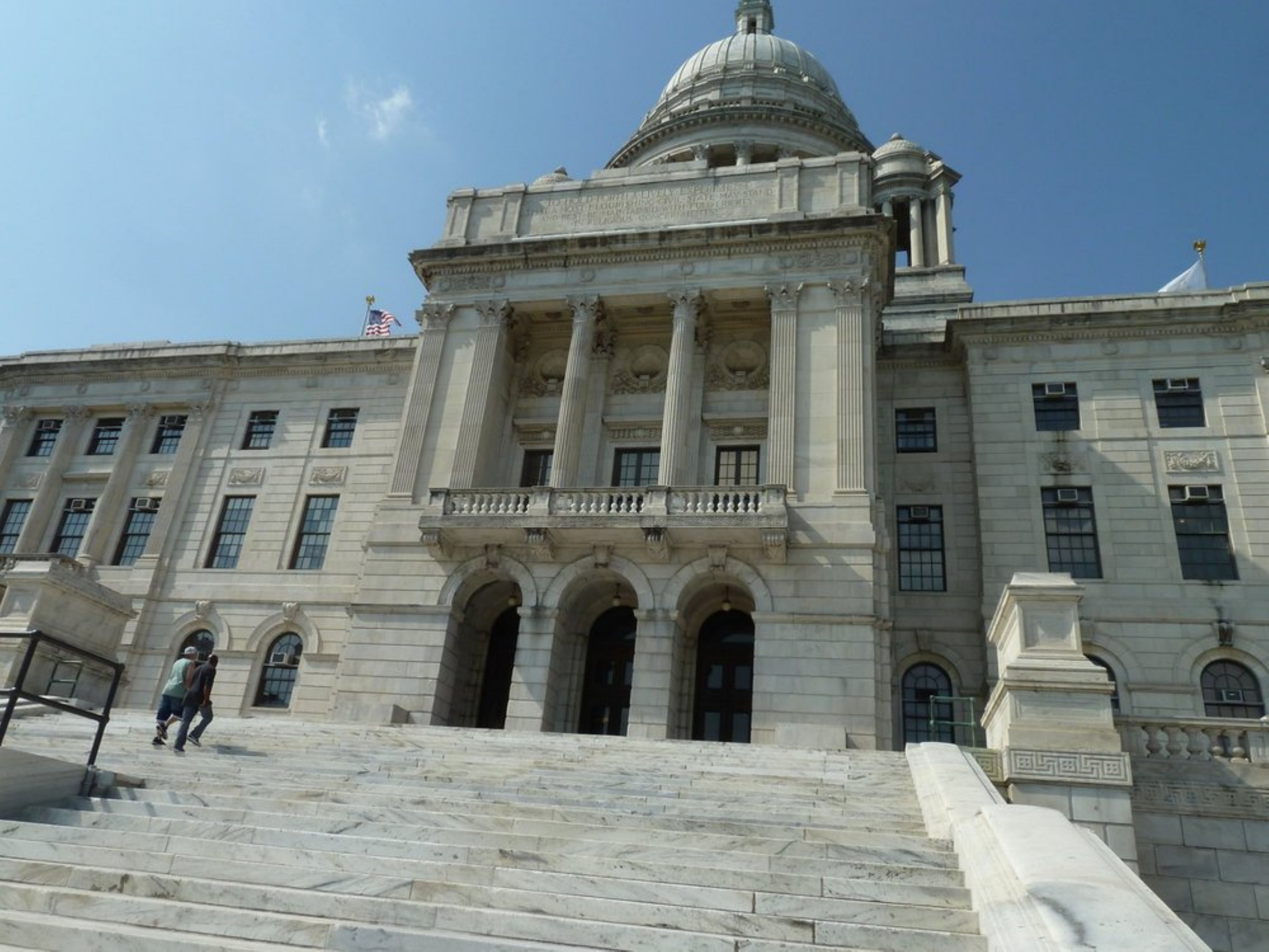}&
\includegraphics[width=0.2\linewidth]{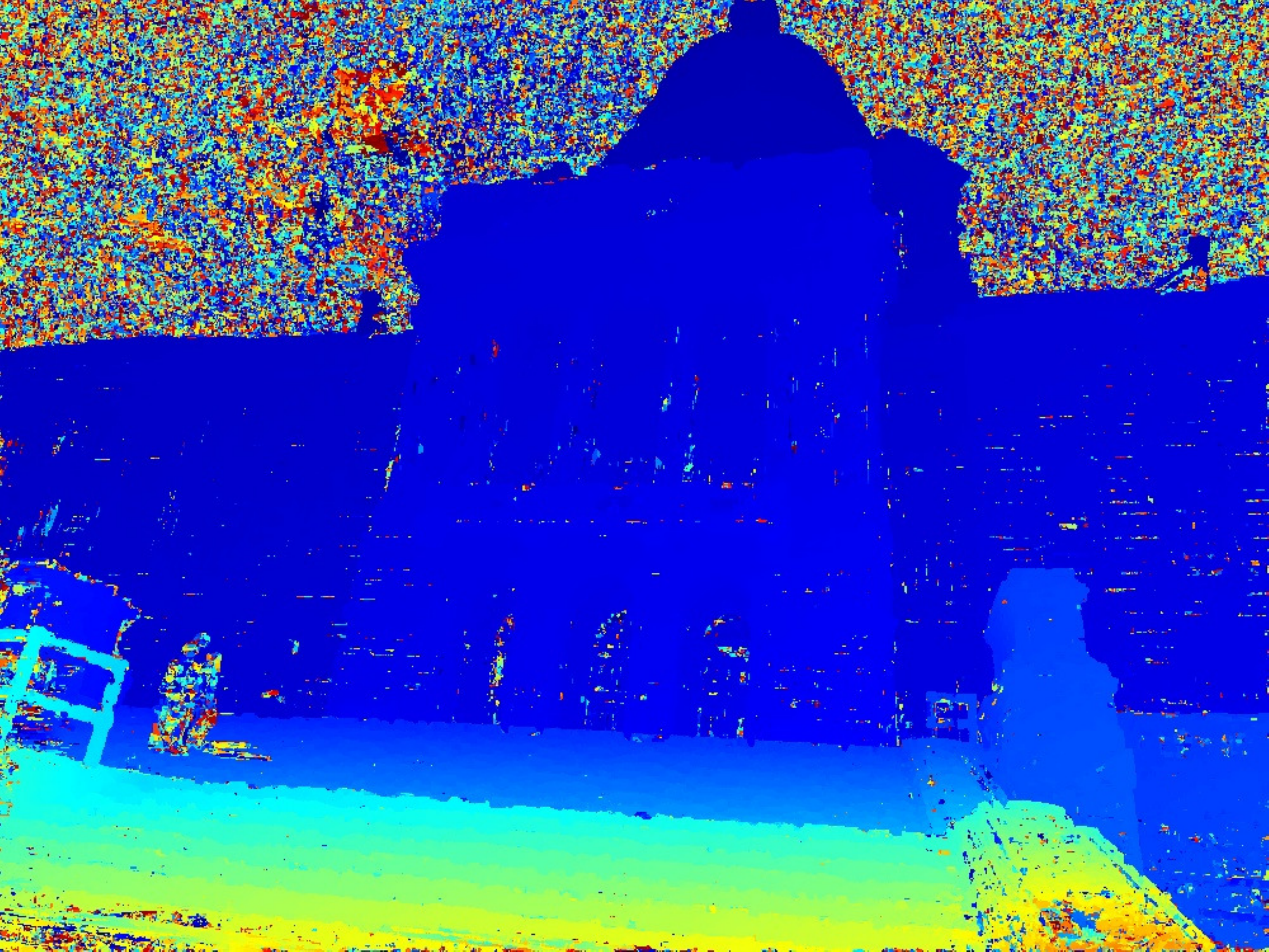}&
\includegraphics[width=0.2\linewidth]{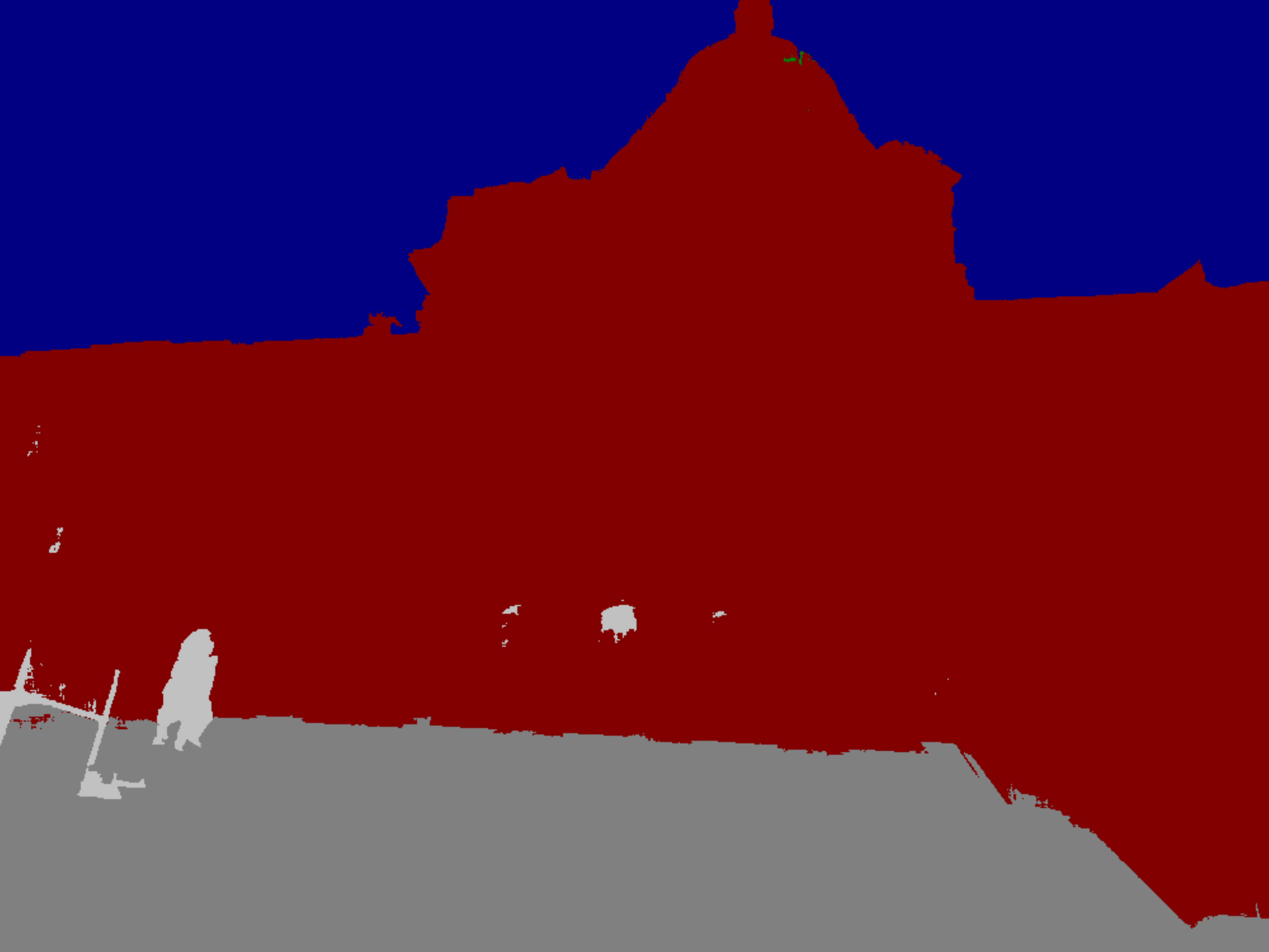}&
\includegraphics[width=0.33\linewidth]{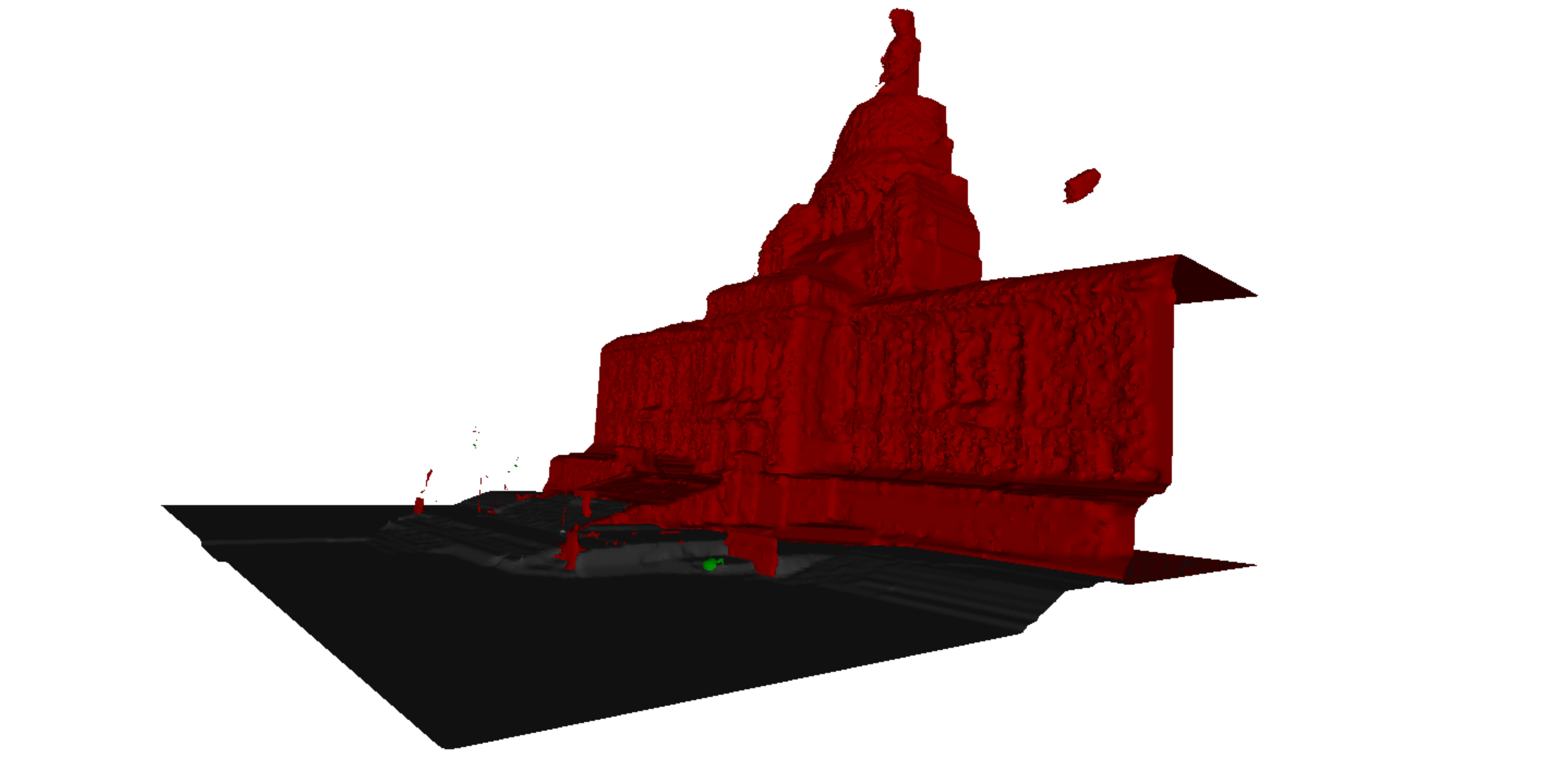}\\
\vspace{4mm}
\includegraphics[width=0.2\linewidth]{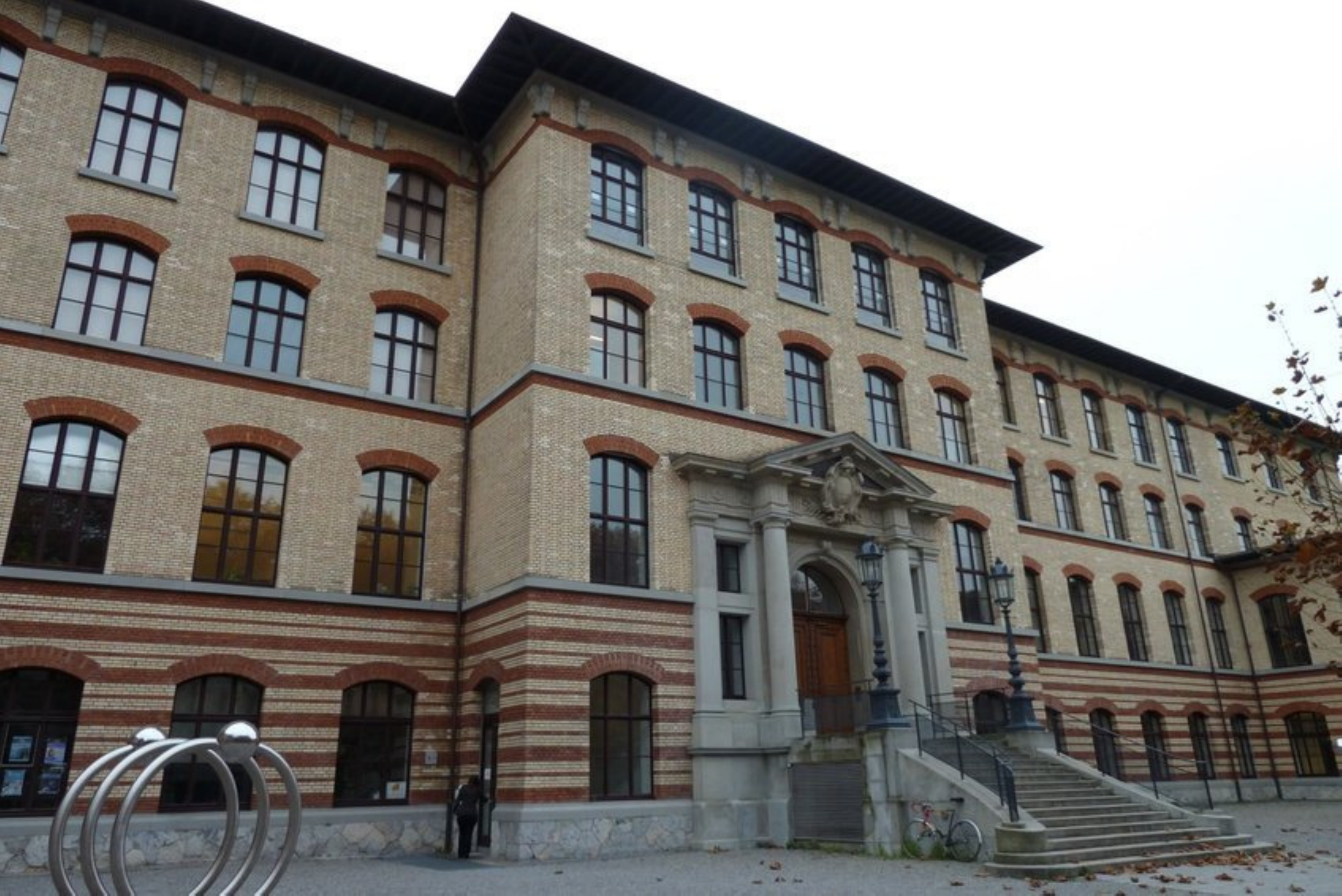}&
\includegraphics[width=0.2\linewidth]{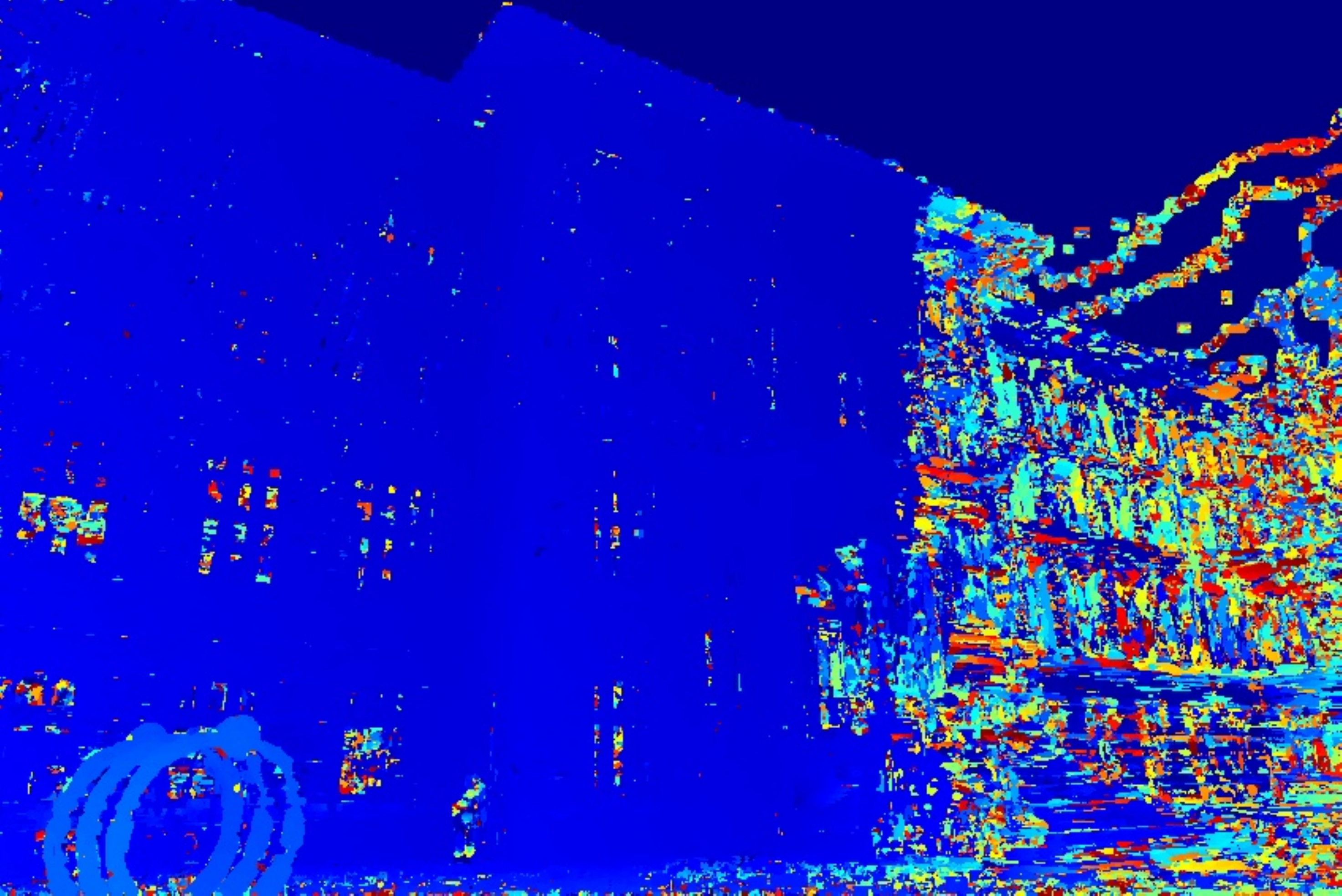}&
\includegraphics[width=0.2\linewidth]{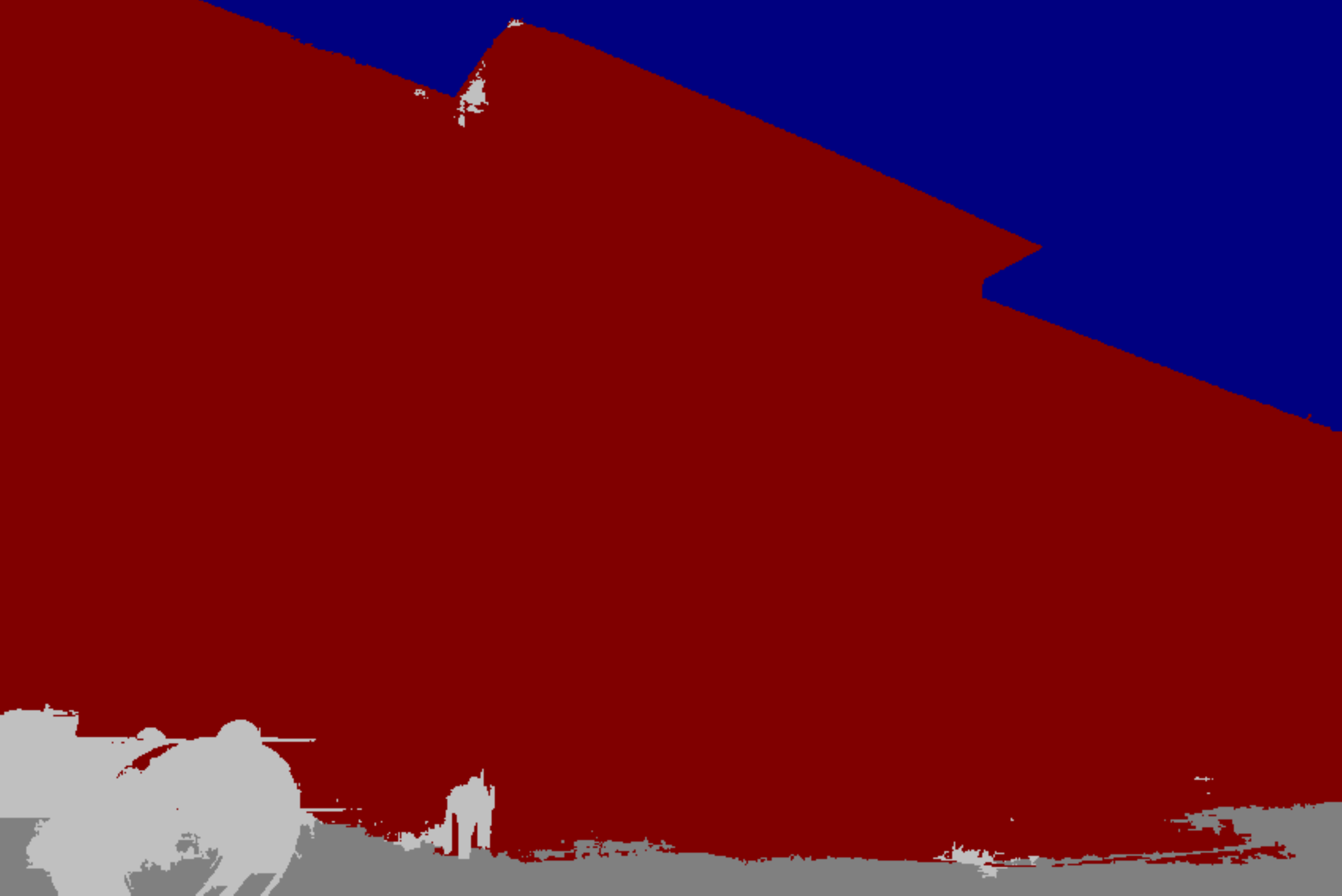}&
\includegraphics[width=0.32\linewidth]{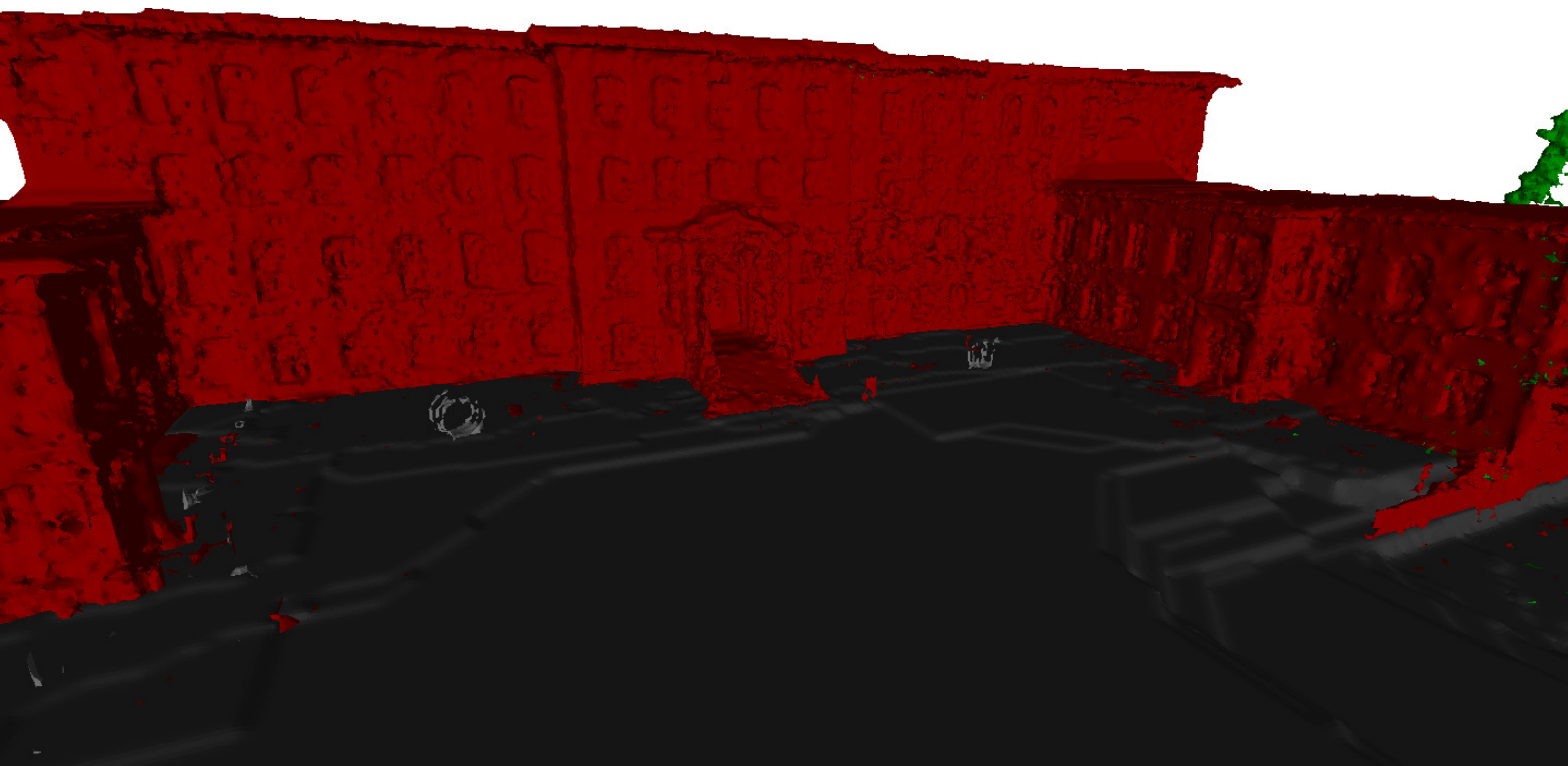}\\
\vspace{4mm}
\includegraphics[width=0.2\linewidth]{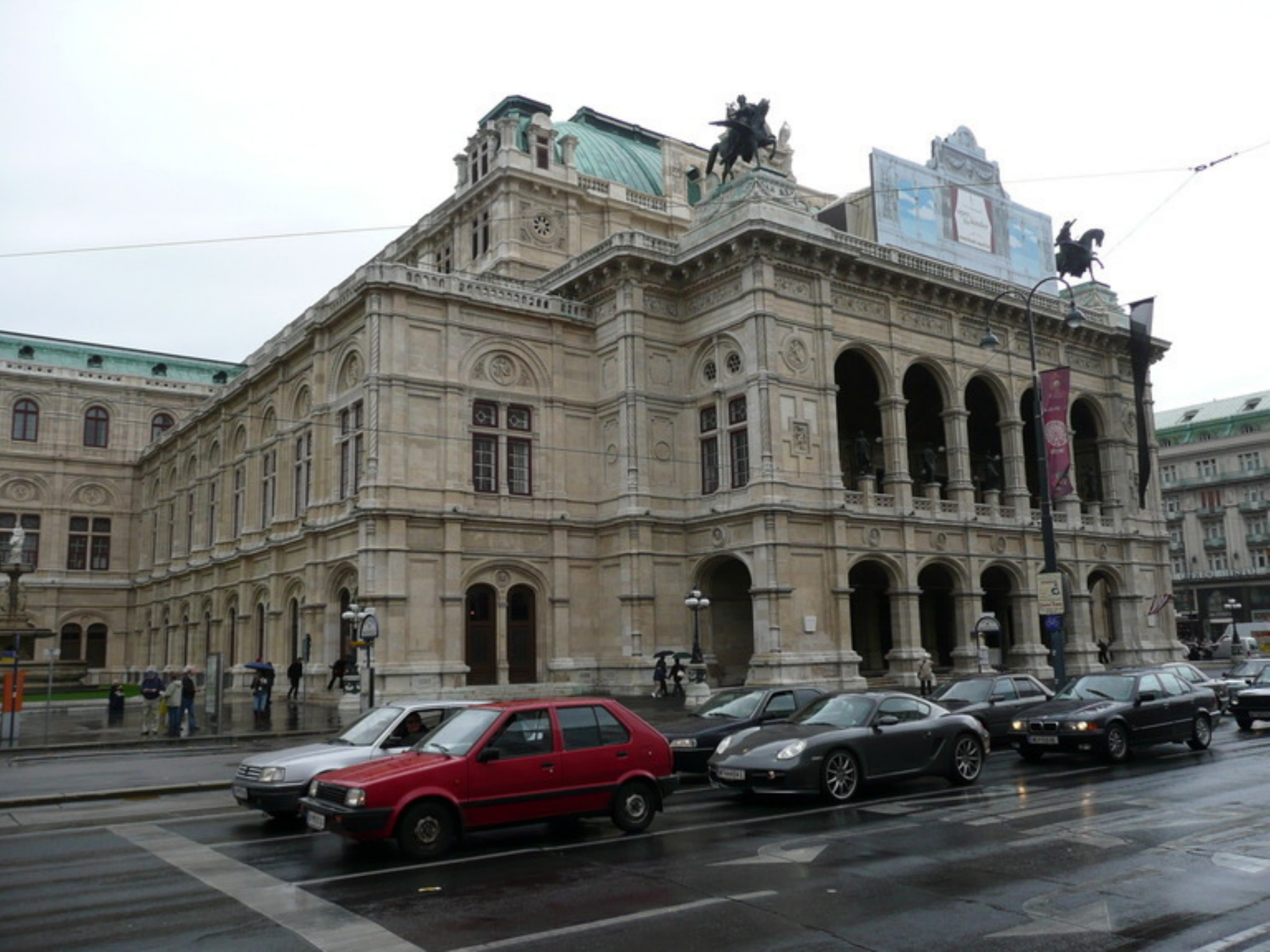}&
\includegraphics[width=0.2\linewidth]{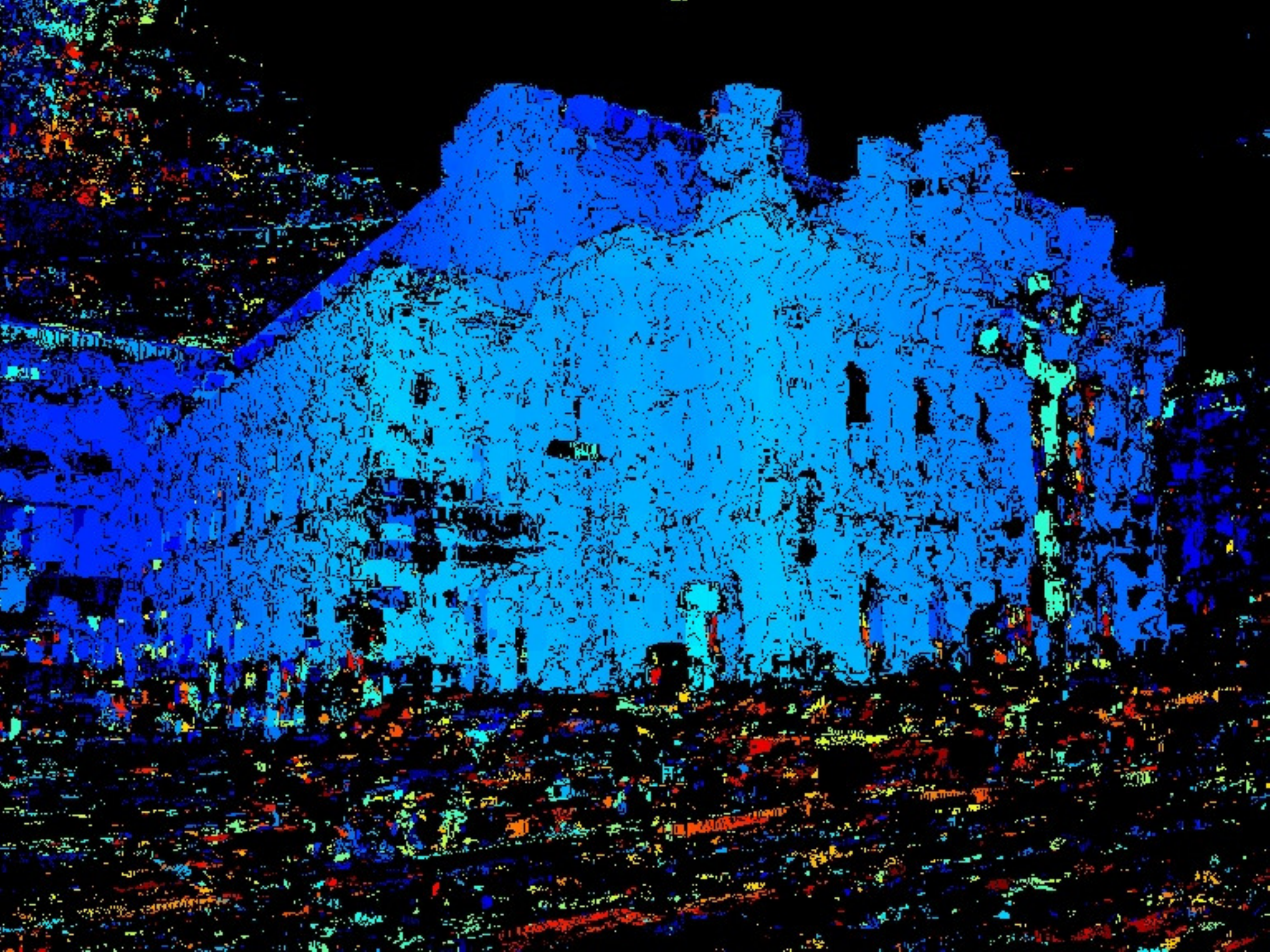}&
\includegraphics[width=0.2\linewidth]{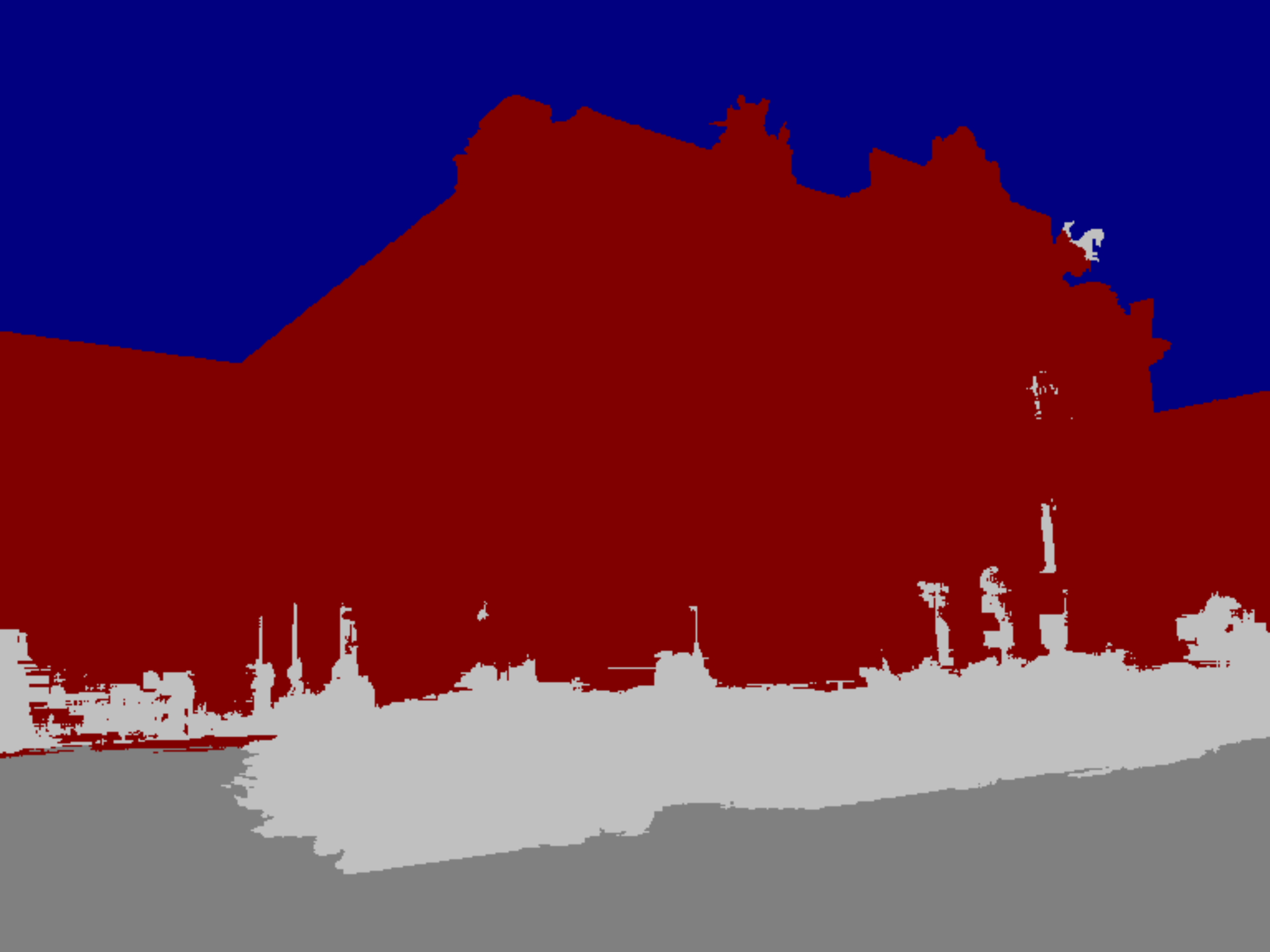}&
\includegraphics[width=0.32\linewidth]{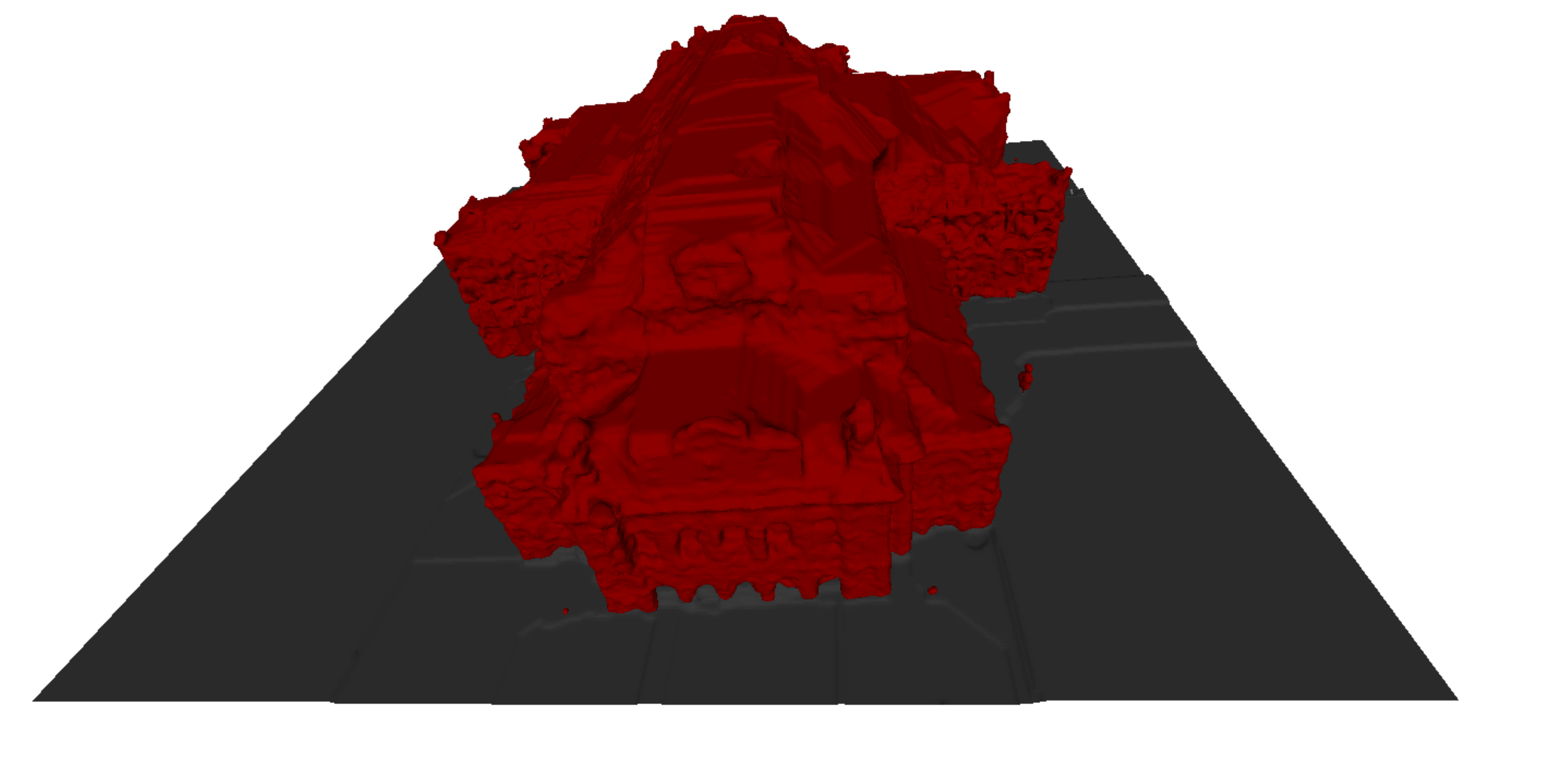}\\
Example Image & Depth Map & Semantic Segmentation & Reconstructed 3D model\\
\end{tabular}
\vspace{2mm}
\caption{\label{fig:results} \it Qualitative results on 6 data sets (from top to bottom): Castle~\cite{Strecha08}, South Building~\cite{Hane13}, Catania~\cite{Hane13}, Providence~\cite{Hane13}, CAB~\cite{Cohen12} and Opera~\cite{Cohen12}. Our method successfully reconstructed challenging 3D data with high level of detail. Minor errors in the reconstructions were caused by the combination of errors of the semantic classifier, insufficient amount data from certain viewpoints or errors in the depth prediction for smooth texture-less surfaces.}
\end{center}
\end{figure*}

\begin{figure*}
\begin{center}
\begin{tabular}{cc}
\includegraphics[width=0.45\linewidth]{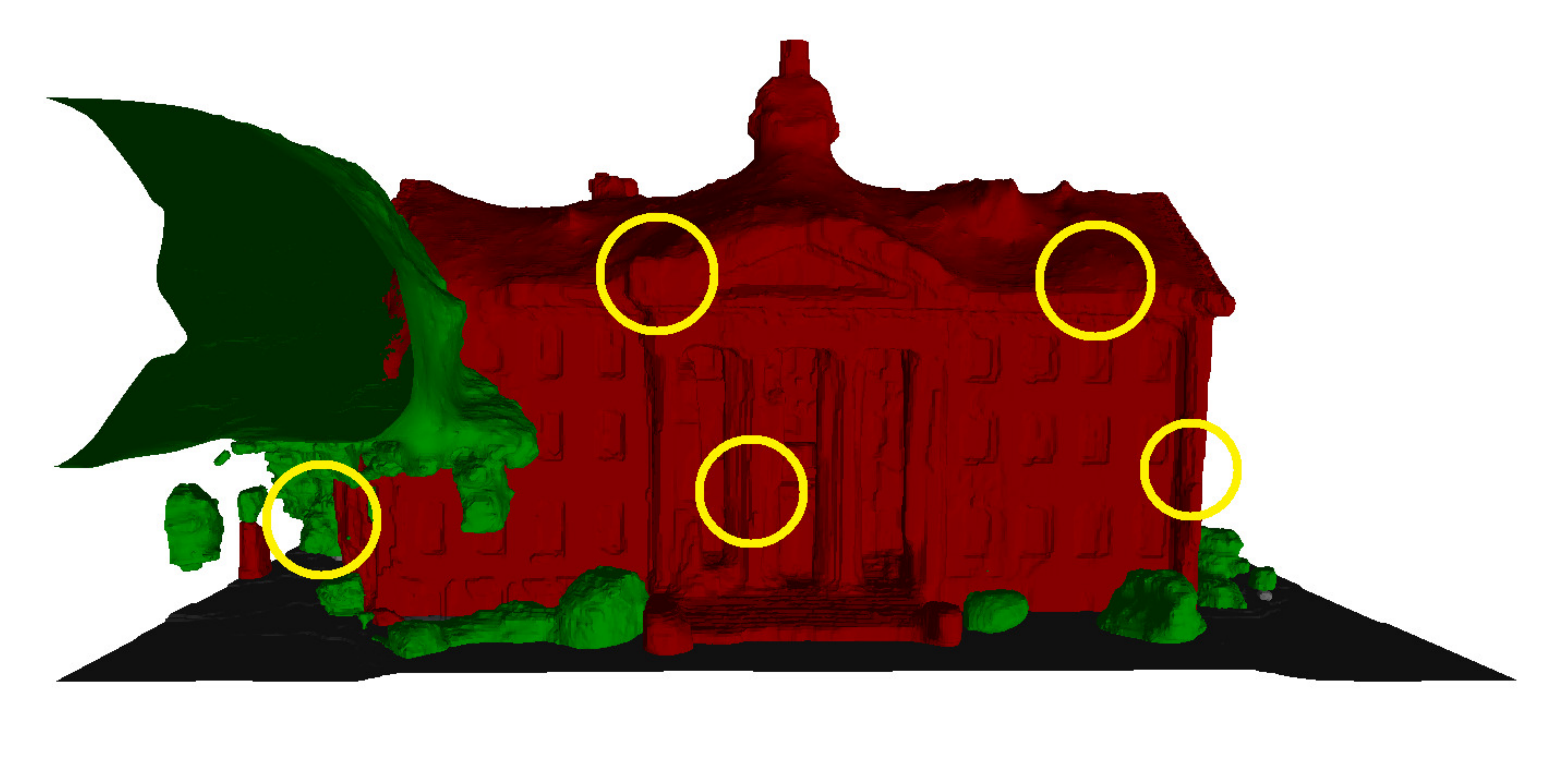}\hspace{9mm}&\hspace{9mm}
\includegraphics[width=0.45\linewidth]{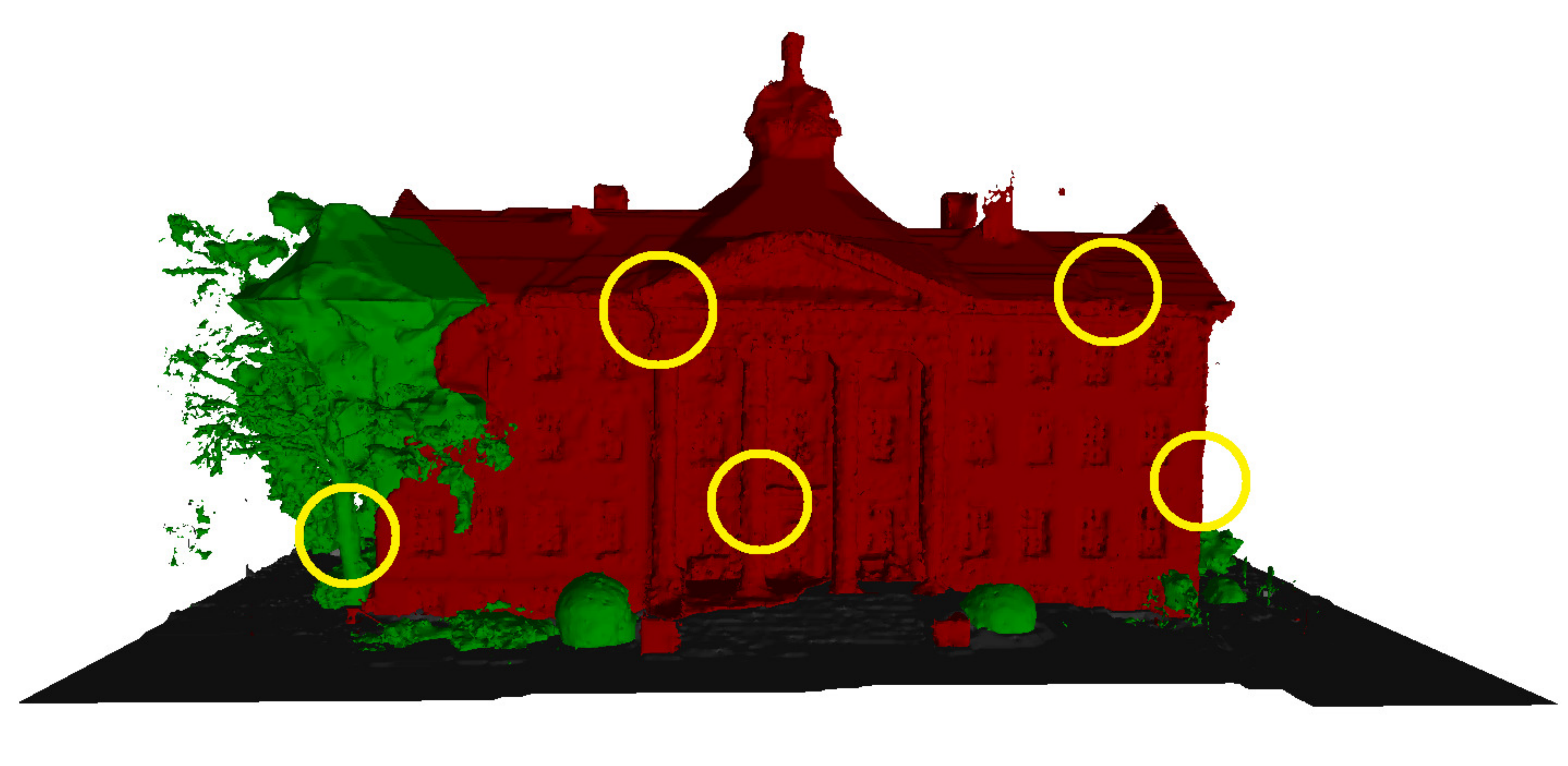}\\
\includegraphics[width=0.45\linewidth]{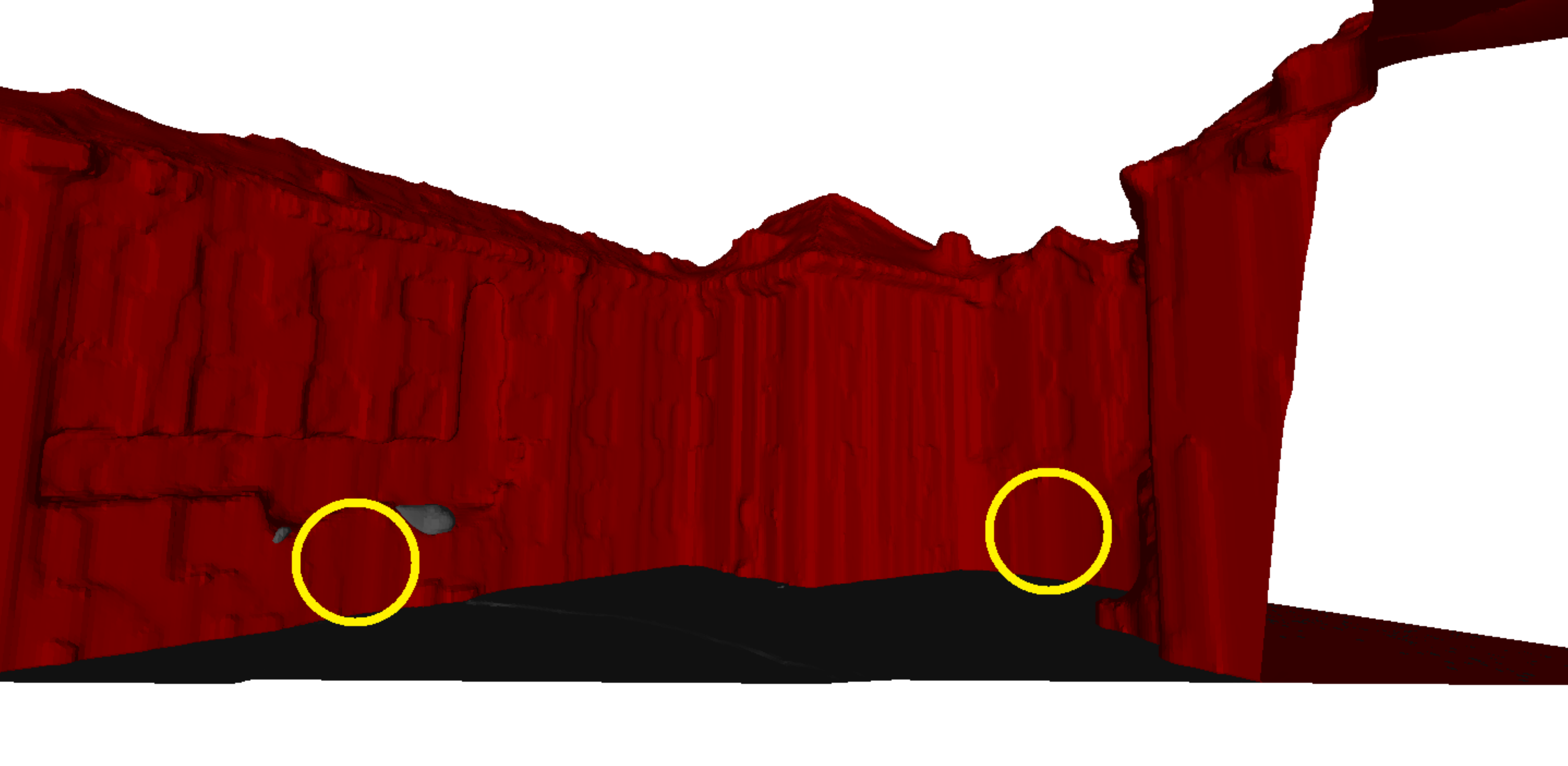}\hspace{9mm}&\hspace{9mm}
\includegraphics[width=0.45\linewidth]{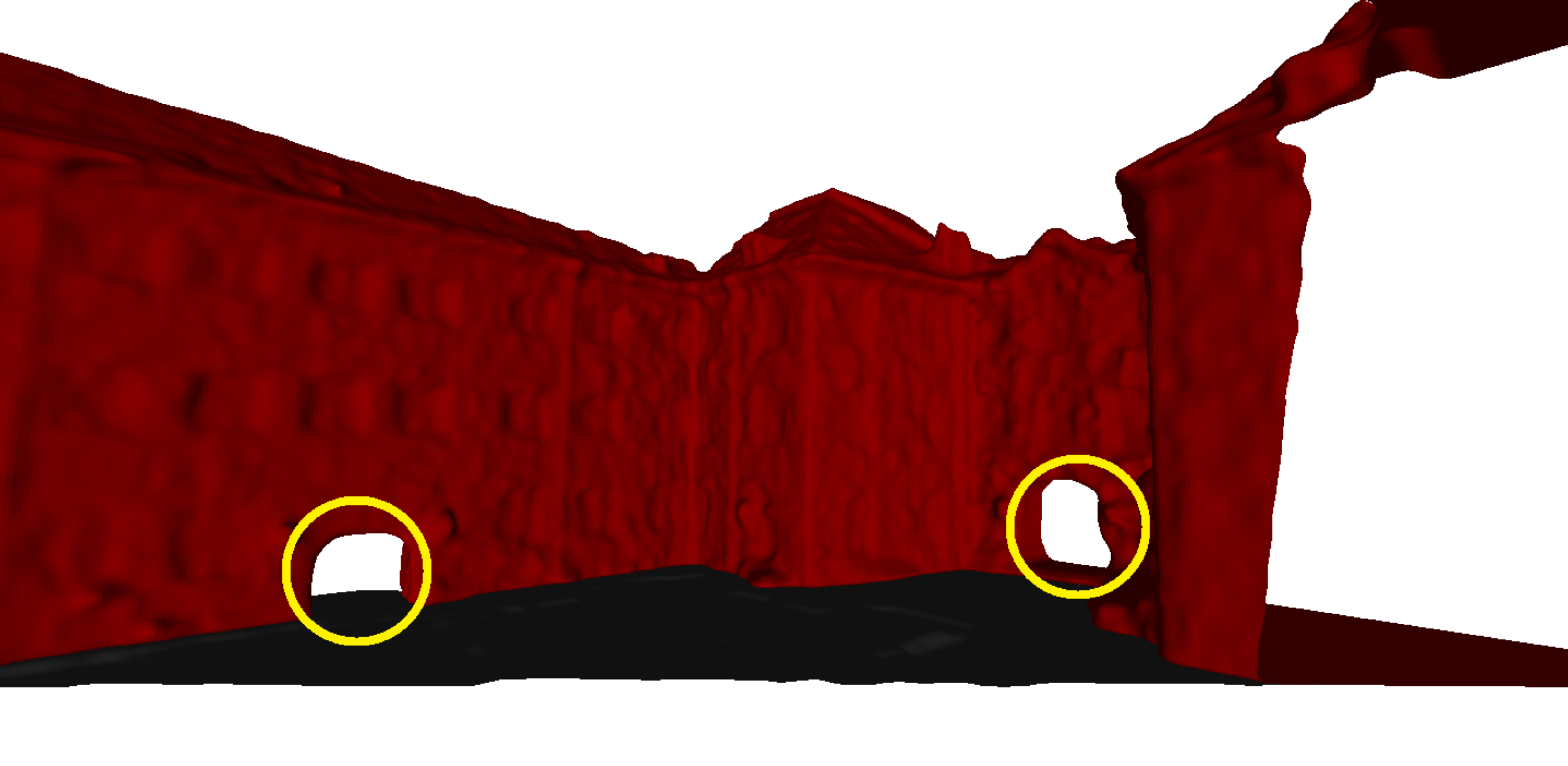}\\
Joint Volumetric Fusion~\cite{Hane13} & Our Method
\end{tabular}
\vspace{2mm}
\caption{\label{fig:comp} \it\small Comparison of method with the sate-of-the-art dense volumetric reconstruction~\cite{Hane13} on South Building~\cite{Hane13} and Castle datasets~\cite{Strecha08}. Incorporation of true semantic and depth ray likelihoods in the optimization framework led to the corrections of systematic reconstruction artifacts, due to the approximations in the formulation of the problem. As can be seen in the South Building reconstruction - state-of-the-art methods, enforcing estimated label in the range behind the depth match, typically causes thickening of the thin structures - columns, tree branches or building corners. Furthermore, as seen in the reconstruction of the Castle, the regularization, minimizing the surface, often closes openings in the flat walls, such as arches in this case, because there is no cost associated. In our framework, such solution has a higher cost, because it disregards otherwise valid matches behind the arch.}
\end{center}
\end{figure*}

We tested our algorithm on $6$ datasets - South Building~\cite{Hane13}, Catania~\cite{Hane13}, CAB~\cite{Cohen12}, Castle-P30~\cite{Strecha08}, Providence~\cite{Hane13} and Vienna Opera~\cite{Cohen12}. The number of images ranged from 30 for Castle to 271 for Opera. The semantic classifier~\cite{hierarchicalcrf} for 5 classes (building, tree, ground, clutter and sky) was trained on the CamVid~\cite{BrostowSFC:eccv08} and MSRC datasets~\cite{ShottonWRC06}, with additional training data from~\cite{Hane13}. Figure~\ref{fig:results} shows the qualitative results for all datasets. Our method managed to successfully reconstruct all 3D scenes with a relatively high precision. Minor problems were caused by insufficient amount of input data and incorrect prediction of semantic and depth estimators. The comparison of models with the state-of-the-art volumetric 3D reconstruction algorithm is shown in the Figure~\ref{fig:comp}. Our method managed to fix systematic reconstruction artifacts, caused by approximations in the modelling of the true ray likelihoods - thin structures tend to be thickened (see columns, roof or tree trunk in the South Building datasets) or openings in the wall (such as arches or doors) undesirably closed, because there is no penalty associated with it.

\section{Conclusion}

In this paper we proposed feasible optimization method for volumetric 3D reconstruction by minimization of reprojection error. Unlike several state-of-the-art methods, our algorithm does not suffer from the systematic errors due to the approximations of corresponding ray potentials. We showed that a direct optimization of the higher order potentials by transformation into pairwise graph under QPBO relaxation is indeed feasible in practice even for high resolution models.  Further work will focus on principled incorporation of other geometric cues in the optimization framework, such as estimated surface normals or planarity enforcing potentials. Another direction would be to investigate the possibility of incorporating reprojection-minimizing ray potentials into other frameworks such as in continuous or mesh-based formulations.

\paragraph{Acknowledgements}
We gratefully acknowledge partial funding by the Swiss National Science Foundation project no. 157101 on 3D image understanding for urban scenes.

{\small
\bibliographystyle{ieee}
\bibliography{ray_cvpr15}
}

\end{document}